\documentclass{article}

\PassOptionsToPackage{numbers, compress}{natbib}

\usepackage[final]{neurips_2019}

\usepackage[utf8]{inputenc} 
\usepackage[T1]{fontenc}    
\usepackage{hyperref}       
\usepackage{url}            
\usepackage{booktabs}       
\usepackage{amsfonts}       
\usepackage{nicefrac}       
\usepackage{microtype}      

\usepackage{wrapfig}
\usepackage{lipsum}
\usepackage[algo2e]{algorithm2e} 
\SetKwInOut{Parameter}{parameter}

\usepackage{color}
\usepackage{tabularx}

\usepackage{algpseudocode}
\usepackage{graphicx}
\usepackage{amsmath}
\usepackage{amssymb}
\usepackage{mathtools}
\usepackage{graphicx}
\usepackage{float} 
\usepackage{subfigure}
\usepackage{algorithm}  
\usepackage{amsmath}  
\usepackage{accents}

\usepackage{comment}
\def\bx{\mathbf x}

\def\by{\mathbf y}

\def\bG{\mathbf G}

\def\mL{\mathcal L}
\def\mD{\mathcal D}
\def\mS{\mathcal S}
\def\btheta{\mathbf \theta}
\def\bTheta{\mathbf \Theta}
\def\btheta{\mathbf \theta}
\def\bTheta{\mathbf \Theta}
\def\T#1{{#1}^{\mathrm{T}}}

\def\mbE{\mathbb{E}}

\def\leref#1{Lemma~\ref{#1}}

\newtheorem{lemma}{Lemma}

\newtheorem{them}{Theorem}

\def\leref#1{Lemma~\ref{#1}}

\def\algref#1{Algorithm~\ref{#1}}

\def\bydef{\triangleq}
\usepackage{amsmath}
\usepackage{accents}

\def\remark{\addtocounter{remark}{1}\def\@currentlabel{\theremark}%
	\emph{Remark~\theremark}. } \makeatother
\newcounter{remark}

\hypersetup{bookmarksnumbered,colorlinks=true}

\title{A Communication-Efficient Collaborative Learning Framework for Distributed Features}

\author{%
  Yang Liu \\
  WeBank\\
  \texttt{yangliu@webank.com} \\
  \And
  Yan Kang \\
  WeBank \\
  \texttt{yangkang@webank.com} \\
  \And
   Xinwei Zhang \\
  University of Minnesota \\
  \texttt{zhan6234@umn.edu} \\
  \And
  Liping Li \\
  University of Science and Technology of China \\
  \texttt{shmily20@mail.ustc.edu.cn} \\
  \And
  Yong Cheng \\
  WeBank \\
  \texttt{petercheng@webank.com}
    \And
  Tianjian Chen \\
  WeBank \\
  \texttt{tobychen@webank.com}
    \And
  Mingyi Hong \\
  University of Minnesota \\
  \texttt{mhong@umn.edu}
      \And
  Qiang Yang \\
  WeBank \\
  \texttt{qyang@cse.ust.hk}
}

\begin{document}

\maketitle
\begin{abstract}
We introduce a collaborative learning framework allowing multiple parties having different sets of attributes about the same user jointly build models without exposing their raw data or model parameters.
In particular, we propose a \textit{Federated Stochastic Block Coordinate Descent (FedBCD)} algorithm, in which each party conducts multiple local updates before each communication to effectively reduce the number of communication rounds among parties, a principal bottleneck for collaborative learning problems. We analyze theoretically the impact of the number of local updates, and show that when the batch size, sample size and the local iterations are selected appropriately, within $T$ iterations, the algorithm performs $\mathcal{O}(\sqrt{T})$ communication rounds and achieves some $\mathcal{O}(1/\sqrt{T})$ accuracy (measured by the average of the gradient norm squared).
The approach is supported by our empirical evaluations on a variety of tasks and datasets, demonstrating advantages over stochastic gradient descent (SGD) approaches.

\end{abstract}

\section{Introduction}\label{sec:introduction}
One critical challenge for applying today's Artificial Intelligence (AI) technologies to real-world applications is the common existence of data silos across different organizations. Due to legal, privacy and other practical constraints, data from different organizations cannot be easily integrated. The implementation of user privacy laws such as GDPR \cite{regulation2016general} has made sharing data among different organizations more challenging. Collaborative learning has emerged to be an attractive solution to the data silo and privacy problem. While distributed learning (DL) frameworks \cite{NIPS2012_4687} originally aims at parallelizing computing power and distributes data identically across multiple servers, federated learning (FL) \cite{DBLP:journals/corr/McMahanMRA16} focuses on data locality, non-IID distribution and privacy. 
In most of the existing collaborative learning frameworks, data are distributed by samples thus share the same set of attributes. 
However, a different scenario is cross-organizational collaborative learning problems where parties share the same users but have different set of features. For example,  a local bank and a local retail company in the same city may have large overlap in user base and it is beneficial for these parties to build collaborative learning models with their respective features. FL is further categorized into horizontal (sample-partitioned) FL, vertical (feature-partitioned) FL and federated transfer learning (FTL) in \cite{DBLP:journals/corr/abs-1902-04885}.

Feature-partitioned collaborative learning problems have been studied in the setting of both DL \cite{Gratton2018,Hu2019FDMLAC,HuADMM} and FL \cite{DBLP:journals/corr/abs-1902-04885,Hardy2017PrivateFL,SecureBoost,ftl}. 
However, existing architectures have not sufficiently addressed the communication problem 
especially in communication-sensitive scenarios where data are geographically distributed and data locality and privacy are of paramount significance (i.e., in a FL setting). In these approaches, \textit{per-iteration} communication and computations are often required, since the update of algorithm parameters needs contributions from all parties. In addition, to prevent data leakage, 
privacy-preserving techniques, such as Homomorphic Encryption (HE) \cite{Rivest1978}, Secure Multi-party Computation (SMPC) \cite{Yao:1982:PSC:1382436.1382751} are typically applied to transmitted data \cite{DBLP:journals/corr/abs-1902-04885,ftl,SecureBoost}, adding expensive communication overhead to the architectures. 
Differential Privacy (DP) is also a commonly-adopted approach, but such approaches suffer from precision loss \cite{Hu2019FDMLAC,HuADMM}. 
In sample-partitioned FL\cite{DBLP:journals/corr/McMahanMRA16}, it is demonstrated experimentally that multiple local updates can be performed with federated averaging (FedAvg), reducing the number of communication round effectively. Whether it is feasible to perform such multiple local update strategy over distributed features is not clear.

In this paper, we propose a collaborative learning framework for distributed features named Federated stochastic block coordinate descent (FedBCD), where parties only share a single value per sample instead of model parameters or raw data for each communication, and can continuously perform local model updates (in either a parallel or sequential manner) without per-iteration communication. In the proposed framework, all raw data and model parameters stay local, and each party does not learn other parties' data or model parameters either before or after the training. There is no loss in performance of the collaborative model as compared to the model trained in a centralized manner. We demonstrate that the communication cost can be significantly reduced by adopting FedBCD. Compared with the existing distributed (stochastic) coordinate descent methods  \cite{CoordinateGD,Mahajan:2017:DBC:3122009.3176835,peng15,niu11}, we show for the first time that when the number of local updates, mini-batch size and learning rates are selected appropriately, the FedBCD converges to a $\mathcal{O}(1/\sqrt{T})$ accuracy with $\mathcal{O}(\sqrt{T})$ rounds of communications despite performing multiple local updates using staled information. 
We then perform comprehensive evaluation of FedBCD against several alternative protocols, including a sequential local update strategy, FedSeq with both the original Stochastic Gradient Descent (SGD) and Proximal gradient descent. 
For evaluation, we apply the proposed algorithms to multiple complex models include logistic regression and convolutional neural networks, and on multiple datasets, including the privacy-sensitive medical dataset MIMIC-III \cite{johnson2016mimic}, MNIST \cite{lecun-mnisthandwrittendigit-2010} and NUS-WIDE \cite{Chua09nus-wide:a} dataset. Finally, we implemented the algorithm for federated transfer learning (FTL) \cite{ftl} to tackle problems with few labeled data and insufficient user overlaps.

\section{Related Work}
Traditional distributed learning adopts a parameter server architecture \cite{NIPS2012_4687} to enable a large amount of computation nodes to train a shared model by aggregating locally-computed updates. The issue of privacy in DL framework is considered in \cite{Shokri:2015:PDL:2810103.2813687}. FL \cite{DBLP:journals/corr/McMahanMRA16} adopted a \texttt{FedAvg} algorithm which runs Stochastic Gradient Descent (SGD) for multiple local updates in parallel to achieve better communication efficiency. 
The authors of~\cite{yu2019parallel} studied the  \texttt{FedAvg}  algorithm under the parallel restarted SGD framework and analyzed the convergence rate and communication savings under IID 
settings.   
In \cite{li2019convergence},   the  convergence of the  \texttt{FedAvg} algorithm under  non-IID settings  was  investigated.
All the work above consider the sample-partitioned scenario.   

Feature-partitioned learning architectures have been developed for models including trees~\cite{SecureBoost}, linear and logistic regression \cite{DBLP:journals/corr/abs-1902-04885,Hardy2017PrivateFL,Gratton2018,HuADMM}, and neural networks \cite{ftl,Hu2019FDMLAC}. Distributed Coordinate Descent \cite{CoordinateGD} used balanced partitions and decoupled computation at the individual coordinate in each partition; Distributed Block Coordinate Descent \cite{Mahajan:2017:DBC:3122009.3176835} assumes feature partitioning is given and performs synchronous block updates of variables which is suited for MapReduce systems with high communication cost settings. These approaches require \textit{synchronization at every iteration}. 
{\it Asynchronous} BCD \cite{peng15} and {\it Asynchronous} ADMM algorithms \cite{niu11} tries to tame various kinds of asynchronicity using strategies such as small stepsize, and careful update scheduling, and the design objective is to ensure that the algorithm can still behave reasonably under non-ideal computing environment.   
Our approach tries to address the expensive communication overhead problem in FL scenario by systematically adopting BCD with sufficient number of local updates guided by theoretical convergence guarantees. 

\section{Problem Definition}
Suppose $K$ data parties collaboratively train a machine learning model based on $N$ data samples  $\{\bx_i, y_i\}_{i=1}^{N}$ and the feature vector $\bx_i \in \mathbb{R}^{1 \times d}$ are distributed among $K$ parties $\{\bx^k_i \in \mathbb{R}^{{1 \times d_k}}\}_{k=1}^{K}$, where $d_k$ is the feature dimension of party $k$. 
Without loss of generality, we assume one party holds the labels and it is party $K$. Let us denote the data set as $\mathcal{D}_i^k\bydef \{\bx_i^k\}$, for $k\in [K-1]$,  $\mathcal{D}_i^K\bydef \{\bx_i^K,y_i^K\}$, and $\mD_i\bydef \{\mD^k_i\}_{k=1}^{K}$ (where $[K-1]$ denotes the set $\{1,\cdots, K-1\}$). 
Then the collaborative training problem can be formulated as 
\begin{align}\label{eq:problem}
\hspace{-0.2cm} & \min_{\bTheta} \mL(\bTheta; \mD)\bydef  \frac{1}{N}\sum^N_{i=1} f(\btheta_1,\dots,\btheta_K; \mD_i)+ \lambda\sum_{k=1}^K\gamma(\theta_k)
\end{align}
where $\theta_k \in \mathbb{R}^{d_k}$ denotes the training parameters of the $k$th party; $\bTheta={[{\btheta_{1}};\dots; {\btheta_{K}}]}$; $f(\cdot)$ and $\gamma(\cdot)$ denotes the loss function and regularizer and $\lambda$ is the hyperparatemer; 
For a wide range of models such as linear and logistic regression, and support vector machines,  the loss function has the following form:
\begin{align}\label{eq:loss_linear}
\hspace{-0.2cm} &  f(\btheta_1,\dots,\btheta_K; \mD_i) = f(\sum_{k=1}^K\bx_i^k\btheta_k,y_i^K)
\end{align}
The objective is for each party $k$ to find its $\btheta_k$ without sharing its data $\mD_i^k$ or parameter $\btheta_k$ to other parties. 

\section{The FedSGD Approach}
If a mini-batch $\mathcal{S}\subset [N]$ of data is sampled, the stochastic partial gradient w.r.t. $\theta_k$ is given by
\begin{equation}\label{eq:sg:1}
g_k(\bTheta; \mS)\bydef\nabla_k f(\bTheta; \mS) + \lambda{\nabla} \gamma(\theta_k).
\end{equation}

Let $H_i^k=\bx_i^k\btheta_k$ and $H_i=\sum_{k=1}^KH_i^k $, then for the loss function in equation (\ref{eq:loss_linear}), we have
\begin{align}\label{grad_linear}
 \nabla_kf(\bTheta; \mS)&= 
 \frac{1}{N_{\mS}}\sum_{i \in \mS}\frac{\partial f(H_i,y_i^K)}{\partial H_i}(\bx_i^k)^T \\
 & \bydef \frac{1}{N_{\mS}}\sum_{i \in \mS}g(H_i,y_i^K)(\bx_i^k)^T
\end{align}
Where $N_{\mS}$ is the number of samples in $\mS$. To compute $ \nabla_k f(\bTheta; \mS)$ 
locally, each party $k \in [K-1]$ sends $ H^k=\{H_i^k\}_{i \in \mS}$ to party $K$, who then calculates $ H^K=\{g(H_i,y_i^K)\}_{i \in \mS}$ and sends to other parties, and finally all parties can compute gradient updates with equation (4). See Algorithm 1.  
\RestyleAlgo{boxruled}
\begin{algorithm}
 \caption{FedSGD implemented on $K$ parties}
\KwIn{learning rate $\eta$}
\KwOut{Model parameters $\theta_1$, $\theta_2$...$\theta_K$}
 Party 1,2..$K$ initialize $\theta_1$, $\theta_2$, ...$\theta_K$.\\
 \For{each iteration j=1,2...}{
{Randomly sample  $S \subset [N]$}\\
{\textbf{Exchange($\{1,2...K\}$)};\\
\For{each party k=1,2..K in parallel}{
    $\boldsymbol k$ computes $g_k$ with equation \eqref{grad1} and update $\theta^{j+1}_k = \theta^{j}_k - \eta g_k$;\\
  }}
}
\textbf{Exchange($U$):} \ \  \textit{\# $U$ is the set of party IDs}  

 \If{equation \eqref{eq:loss_linear} hold}{
    \For{each party $k \in U$ and $k \ne\ K$ in parallel}{
     $\boldsymbol k$ computes $H^k$,and send $H^k$ to party $K$;\\}
     party $K$ computes $H^K$ and sends to all other parties in $U$;\\}
\Else{ 
    \For{each party $k \in U$ in parallel}{
     $\boldsymbol k$ computes $H_k^1$...$H_k^K$,and send $H_k^q$ to party $q$;\\}} 
\label{algo1}
\end{algorithm}

For an arbitrary loss function, let us define the collection of information needed to compute $\nabla_k f(\bTheta; \mS)$ as
\begin{align}
H^{S}_{-k}:=\{\it{H}_q^k(\theta_q,\mathcal{S}^q)\}_{q\ne k}. 
\end{align}
where $\it{H}_q^k(\cdot)$ is a function summarizing the information required from party $q$ to $k$, the stochastic gradients  \eqref{eq:sg:1} can be computed as the following:
\begin{align}\label{grad1}
g_k(\bTheta; \mS)&= \nabla_k f(H^{S}_{-k}, \theta_k; \mS) + \lambda{\nabla} \gamma(\theta_k)\nonumber\\
& \bydef  g_k(H_{-k}, \theta_k; \mS).
\end{align}

Therefore, the overall stochastic gradient is given as
\begin{align}
    g(\bTheta; \mS) \triangleq [g_1(H_{-1},\theta_1; \mS);\cdots; g_K(H_{-K}, \theta_K;\mS)].
\end{align}


A direct approach to optimize \eqref{eq:problem} is to use the vanilla stochastic gradient descent (SGD) algorithm given below
\begin{equation}\label{grad0}
\theta_k  \leftarrow \theta_k - \eta g_k(H_{-k},\theta_k; \mS),\quad \forall~k.
\end{equation}
The federated implementation of the above SGD iteration is given in Algorithm \ref{algo1}.  Algorithm \ref{algo1} requires communication of intermediate results {\it at every iteration}. This could be very inefficient, especially when $K$ is large or the task is communication heavy. For a task of form in equation (\ref{eq:loss_linear}), the number of communications per round is $2(K-1)$, but for an arbitrary task, the number of communications per round can be $K^2-K$ if it requires pair-wise communication. 
We note that since Algorithm 1 has the same iteration as the vanilla SGD algorithm, it converges with a rate of  $\mathcal{O}(\frac{1}{\sqrt{T}})$, regardless of the choice of $K$ \cite{ghadimi2013stochastic}. Since each iteration requires one round of communication among all the parties, $T$ rounds of communication is required to achieve an error of $\mathcal{O}(\frac{1}{\sqrt{T}})$.

\section{The Proposed FedBCD Algorithms}
In the parallel version of our proposed algorithm, called \textit{FedBCD-p}, at each iteration, each party $k\in [K]$ performs $Q>1$ consecutive local gradient updates {\it in parallel}, before communicating the intermediate results among each other; see Algorithm \ref{algo2}. Such ``multi-local-step" strategy is strongly motivated by our practical implementation (to be shown in our Experiments Section), where we found that performing multiple local steps can significantly reduce overall communication cost. Further, such a strategy also resembles the \texttt{FedAvg} algorithm \cite{DBLP:journals/corr/McMahanMRA16}, where each party performs multiple local steps before aggregation.

\RestyleAlgo{boxruled}
\begin{algorithm}
	\caption{FedBCD-p: Parallel  Federated Stochastic Block Coordinate Descent}
	\KwIn{learning rate $\eta$}
	\KwOut{Model parameters $\theta_1$, $\theta_2$...$\theta_K$}
	Party 1,2..$K$ initialize $\theta_1$, $\theta_2$, ...$\theta_K$.\\
	\For{each outer iteration t=1,2...}{
		{Randomly sample a mini-batch $\mS\subset\mD$};\\
		{
			\textbf{Exchange ($\{1,2...K\}$)};\\
			\For{each party $k\in [N]$, in parallel}{
				\For{each local iteration $j=1,2...,Q$}{
					$\boldsymbol k$ computes $g_k(H_{-k}, \theta_k;\mS)$ using \eqref{grad1};\\  
					Update $\theta_k \leftarrow \theta_k - \eta { g_k(H_{-k},\theta_k;\mS)}$;
				}}
			}
		}
		\label{algo2}
	\end{algorithm}
	
At each iteration the $k$th feature is updated using the direction (where $\mS$ is a mini-batch of data points)
\begin{align}
    d_k = -\eta g_k(H_{-k},\theta_k;\mS).
\end{align}
Because $H_{-k}$ is the intermediate information obtained from the {\it most recent synchronization}, it may contain {\it staled} information so it may no longer be an unbiased estimate of the true partial gradient $\nabla_k \mathcal{L}(\bTheta)$. 
On the other hand, during the $Q$ local updates no inter-party communication is required. Therefore, one could expect that there will be some interesting tradeoff between communication efficiency and computational efficiency. In the same spirit, a sequential version of the algorithm allows the parties to update their local $\theta_k$'s sequentially, while each update consists of $Q$ local updates without inter-party communication, termed \textit{FedBCD-s}. (Algorithm \ref{algoseq}).

\RestyleAlgo{boxruled}
\begin{algorithm}
 \caption{Sequential FedBCD: Sequential  Federated Stochastic Block Coordinate Descent}
\KwIn{learning rate $\eta$}
\KwOut{Model parameters $\theta_1$, $\theta_2$...$\theta_K$}
 Party 1,2..$K$ initialize $\theta_1$, $\theta_2$, ...$\theta_K$.\\
 \textbf{Exchange($\{1,2...K\}$)};\\
 \For{each iteration i=1,2...}{
Randomly sample a mini-batch $\mS\subset\mD$;\\
{
 \For{each party k=1,2..K sequentially}{
  \For{each local iteration j=1,2...,Q}{ 
  $\boldsymbol k$ computes $g_k$ using \eqref{grad1} and update $\theta_k \leftarrow \theta_k - \eta g_k(H_{-k},\theta_k;\mS)$;\\
  }
  \textbf{Exchange($\{k,K\}$)};\\
  }
  }
  }
 
\label{algoseq}
\end{algorithm}

\section{Security Analysis} 

Here we aim to find out whether one party can learn other party's data from collections of messages exchanged ($H_q^k$) during training. Whereas previous research studied data leakage from exposing complete set of model parameters or gradients~\cite{ZhuNIPS20199617,Hitaj2017,Melis2018ExploitingUF}, in our protocol model parameters are kept private, and only the intermediate results (such as inner product of model parameters and feature) are exposed.

\paragraph{Security Definition} 

Let $\mS_{j}$ be the set of data point sampled at the $j$th iteration and $i_j$ denotes the $i$th sample of the $j$th iteration. $H^k_{i_j}$ is the contribution of the $i$th sample to other parties. At the $(j+1)$th iteration, we update weight variables according to equation (\ref{grad_linear})

\begin{equation}\label{iter_grad}
    \theta^{j+1}_k =\theta^{j}_k -\eta_j (\frac{1}{N_{\mS_j}}\sum_{i_j\in\mS_j}g(H_{i_j},y^K_{i_j})(\bx_{i_j}^k)^T+\lambda \theta_k^{j})
\end{equation}

The security definition is that for any party $k$ with undisclosed dataset $\mD^k$ and training parameters $\theta_k$ following FedBCD, there exists infinite solutions for $\{x_{i_j}^k\}_{i \in \mS_{j,j=0,1\cdots,M}}$ that yield the same set of contributions $\{H^k_{{i_j}}\}_{j=0,1\cdots,M}$. That is, \textit{one can not determine party $k$'s data uniquely from its exchanged messages of $\{H^k_{{j}}\}_{j=0,1\cdots,M}$ regardless the value of $M$, the total number of iterations}.

Such a security definition is inline with prior security definitions proposed in privacy-preserving machine learning and secure multiple computation (SMC), such as \cite{bingsheng_aaai,79888e1f5b584e36a5fcbaea9e60585e,Vaidya:2002:PPA:775047.775142,DBLP:journals/corr/abs-1902-04885,Mangasarian:2008:PCV:1409620.1409622}. Under this security definition, when some prior knowledge about the data is known, an adversary may be able to eliminate some alternative solutions or certain derived statistical information may be revealed \cite{79888e1f5b584e36a5fcbaea9e60585e,Vaidya:2002:PPA:775047.775142} but it is still impossible to infer the exact raw data ("deep leakage"). However, this practical and heuristic security model provides flexible tradeoff between privacy and efficiency and allows much more efficient solutions. Our problem is particularly challenging in that the observations by other parties \textit{are serial outputs from FedBCD algorithm and are all correlated based on equation (\ref{iter_grad})}. Although it is easy to show security to send one round of $H^k_{i_j}$ due to protection from the reduced dimensionality, it is unclear whether raw data will be leaked after thousands or millions of rounds of iterative communications. 

\begin{them}\label{th:security}
For $K$-party collaborative learning framework following \eqref{eq:loss_linear} with $K\geq 2$, the FedBCD Algorithm is secured for party $k$ if $k$'s feature dimension is larger than 2, i.e., $d_k \geq 2$.
\end{them}

\begin{figure}[tb!]
    \centering
    \subfigure[]{\includegraphics[width=0.45\linewidth,height=3cm]{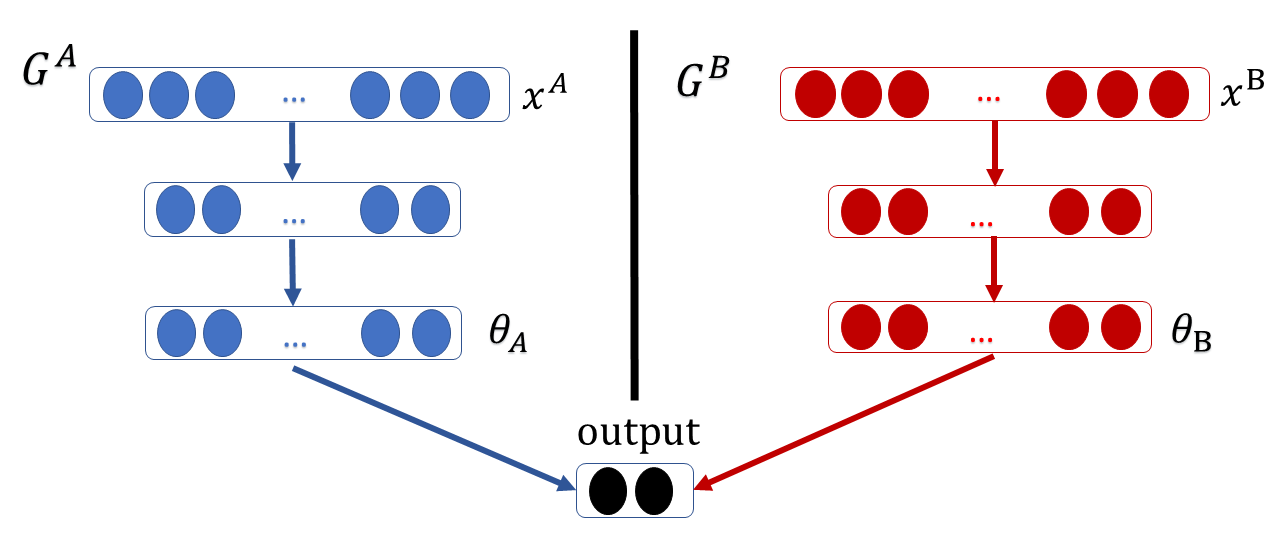}
    \label{submodel}}
    \subfigure[]{\includegraphics[width=0.45\linewidth,height=3cm]{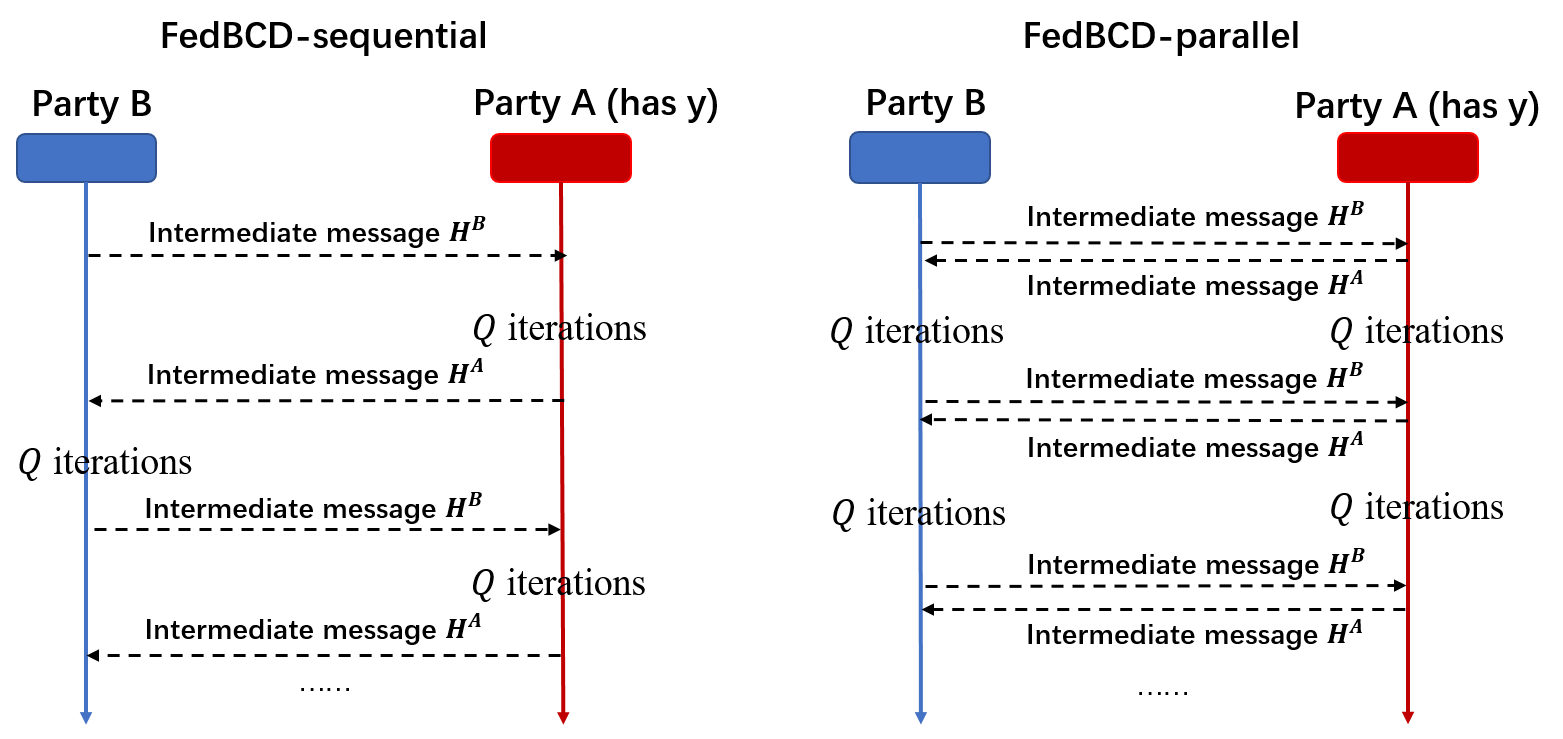}
    \label{fedbcd_sq}}
    \caption{Illustration of a 2-party collaborative learning framework (a) with neural network(NN)-based local model. (b) \textit{FedBCD-s} and \textit{FedBCD-p} algorithms}
    \label{nn_fl}
\end{figure}

The security proof can be readily extended to collaborative systems where parties have arbitrary local sub-models (such as neural networks) but connect at the final prediction layer with loss function (\ref{eq:loss_linear}) (see Figure \ref{submodel}). Let $G^k$ be the local transformation on $\bx_i^k$ and is unknown to parties other than $k$. We choose $G_i^k$ to be the identity map, i.e. $G_i^k(x_i^k)=(x_i^k)$, then the problem reduces to Theorem \ref{th:security}.

\section{Convergence Analysis}
In this section, we perform convergence analysis of the FedBCD algorithm. Our analysis will be focused on the parallel version Algorithm 2 and the sequential version can be analyzed use similar techniques. 

To facilitate the proof, we define the following notations: Let $r$ denote the iteration index, in which each iteration one round of local update is performed; Let $r_0$ denote the latest iteration before $r$ in which synchronization has been performed, and the intermediate information $H^k$'s are exchanged. Let
$\by^r_k$ denote the ``local vector" that node $k$ uses to compute its local gradient at iteration $r$, that is
\begin{align} \label{ca_local_grad}
   g_k(\by^r_k;\mS)= g_k([\bTheta^{r_0}_{-k},\theta^r_k];\mS])
\end{align}
where $[\mathbf{v}_{-k},w]$ denotes a vector $\mathbf{v}$ with its $k$th element replaced by $w$. Note that by Algorithm 2, each node $k$ always updates the $k$th element of $\by^r_k$, while the information about  $\bTheta^{r_0}_{-k}$ is obtained by the most recent synchronization step. 
Further, we use the ``global" variable $\bTheta^r$ 
to collect the most updated parameters at each iteration of each node, where $\by_{k,j}$ denotes the $j$th element of $\by_k$:
$$\bTheta^r={[{\btheta^{r}_{1}};\dots; {\btheta^{r}_{K}}]}\bydef{[{\by^{r}_{1,1}};\dots;{\by^{r}_{K,K}}]}.$$  
Note that $\{\bTheta^r\}$ is only a sequence of \textquotedblleft virtual" variables, it is never explicitly formed in the algorithm.

\noindent\textbf{Assumptions} 

\noindent\textbf{A1:} {\bf Lipschitz Gradient}. Assume that the loss function satisfies the following: 
\begin{align*}
&\lVert \nabla \mL(\bTheta_1)-\nabla \mL(\bTheta_2)\rVert\leq L\lVert \bTheta_1-\bTheta_2\rVert, \; \forall~\bTheta_1, \bTheta_2\\
&\lVert \nabla_k \mL(\bTheta_1)-\nabla_k \mL(\bTheta_2)\rVert\leq L_k\lVert \bTheta_1-\bTheta_2\rVert, \; \forall~\bTheta_1, \bTheta_2.
\end{align*}

\noindent\textbf{A2:} {\bf Uniform Sampling}. For simplicity, assume that the data sample is partitioned into $B$ mini-batches $\mS_1,\cdots, \mS_B$, each with size $S$; at a given iteration, $\mS$ is  sampled uniformly from these mini-batches. 

Our main result is shown below. The detailed analysis is relegated to the Supplemental Material. 

\begin{them}\label{th:main}
	Under assumption A1, A2, when the step size in Algorithm 2 satisfies $0<\eta\leq \min\{\frac{\sqrt{2}}{2Q\sqrt{\sum^K_{j=1}L^2_j+3L^2_k}},\frac{1}{L}\}$, then for all $T\geq1$, we have the following bound:
	\begin{align}\label{eq:thm}
	\frac{1}{T}\sum^{T-1}_{r=0}&\mbE[\lVert \nabla\mL(\bTheta^r)\rVert^2]\leq\frac{2}{\eta T}(\mL(\bTheta^{(0)})-\mL(\bTheta^{\star}))  \\
	& +2\eta^2(K+3)Q^2\sum^{K}_{k=1}L^2_k\frac{\sigma^2}{S}+2K\frac{\sigma^2}{S}.
	\end{align}
	where $\mL(\bTheta^{\star})$ denotes the global minimum of problem~\eqref{eq:problem}.
\end{them}

\remark{
It is non-trivial to find an unbiased estimator for the local stochastic gradient $g_k(\by^r_k;\mS)$. This is because after each synchronization step, each agent $k$ performs $Q$ {\it deterministic steps} based on the same data set $\mS$, while fixing all the rest of the variable blocks at $\bTheta^{r_0}_{-k}$. This is significantly different from FedAvg-type algorithms, where at each iteration a new mini-batch is sampled at each node.}


\remark{If we pick $\eta=\frac{1}{\sqrt{T}}$, and $S=Q=\sqrt{T}$, then with any fixed $K$ the convergence speed of the algorithm is $\mathcal{O}(\frac{1}{\sqrt{T}})$ (in terms of the speed of shrinking the averaged gradient norm squared). 
To the best of our knowledge, it is the first time that such an $\mathcal{O}(1/\sqrt{T})$ rate has been proven for any algorithms with multiple local steps designed for the feature-partitioned collaboratively learning problem.}  

\remark{Compared with the existing distributed stochastic coordinate descent methods  \cite{CoordinateGD,Mahajan:2017:DBC:3122009.3176835,peng15,niu11}, our results are different. It shows that, despite using stochastic gradients and performing multiple local updates using staled information, only $\mathcal{O}(\sqrt{T})$ communication rounds are requires (out of total $T$ iterations) to achieves  $\mathcal{O}(1/\sqrt{T})$ rate. We are not aware of any of such guarantees for other distributed coordinate descent methods.}

\remark{If we consider the impact of the number of nodes $K$ and pick $\eta=\frac{1}{\sqrt{KT}}$, and $S=Q=\sqrt{TK}$, then the convergence speed of the algorithm is $\mathcal{O}(\frac{\sqrt{K}}{\sqrt{T}})$. This indicates that the proposed algorithm has a  slow down w.r.t the number of parties involved. However, in practice, this factor is mild assuming that the total number of parties involved in a feature-partitioned collaboratively learning problem is usually not large.}

\section{Experiments}
\noindent\textbf{Datasets and Models}

\noindent\textbf{MIMIC-III.} MIMIC-III (Medical Information Mart for Intensive Care) \cite{johnson2016mimic} is a large database comprising information related to patients admitted to critical care units at a large tertiary care hospital. Data includes
vital signs, medications, survival data, and
more. Following the data processing procedures of \cite{harutyunyan2017multitask}, we compile a subset of the MIMIC-III database containing more than 31 million clinical events that correspond to 17 clinical variables and get the final training and test sets of 17,903 and 3,236 ICU stays, respectively. For each variable we compute six different sample statistic features on seven different subsequences of a given time series, obtaining $17 \times 7 \times 6=714$ features. We focus on the in-hospital mortality prediction task based on the first 48 hours of an ICU stay with area under the receiver operating characteristic (AUC-ROC) being the main metric. We partition each sample vertically by its clinical features. 
In a practical situation, clinical variables may come from different hospitals or different departments in the same hospital and can not share their data due to the patients personal privacy. This task is refered to as MIMIC-LR.

\noindent\textbf{MNIST.} We partition each MNIST ~\cite{lecun-mnisthandwrittendigit-2010} image with shape $28 \times 28 \times 1$ vertically into two parts (each part has shape $28 \times 14 \times 1$). Each party uses a local CNN model to learn feature representation from raw image input. The local CNN model for each party consists of two 3x3 convolution layers that each has 64 channels and are followed by a fully connected layer with 256 units. Then, the two feature representations produced by the two local CNN models respectively are fed into the logistic regression model with 512 parameters for a binary classification task. We refer this task as MNIST-CNN. 

\noindent\textbf{NUS-WIDE.} The NUS-WIDE dataset \cite{Chua09nus-wide:a} consists of 634 low-level images features extracted from Flickr images as well as their associated tags and ground truth labels. We put 634 low-level image features on party B and 1000 textual tag features with ground truth labels on party A. The objective is to perform a federated transfer learning (FTL) task studied in  \cite{ftl}. FTL aims to predict labels to unlabeled images of party B through transfer learning form A to B. Each party utilizes a neural network having one hidden layer with 64 units to learn feature representation from their raw inputs. Then, the feature representations of both sides are fed into the final federated layer to perform federated transfer learning. This task is refered to as NUS-FTL.

\noindent\textbf{Default-Credit.} The Default-Credit consists of credit card records including user demographics, history of payments, and bill statements, etc., with a user's default payments as labels. 
We separate the features in a way that the demographic features are on one side, separated from the payment and balance features. This segregation is common in industry applications such as retail and car rental leveraging banking data for user credibility prediction and customer segmentation. In our experiments, party A has labels and 18 features including six months of payment and bill balance data, whereas party B has 15 features including user profile data such as education, marriage 
. We perform a FTL task as described above but with homomorphic encryption applied. We refer to this task as Credit-FTL.

For all experiments, we adopt a decay learning rate strategy with $\eta^t= \frac{\eta0}{\sqrt{t+1}}$, where $\eta0$ is optimized for each experiment. We fix the batch size to 64 and 256 for MIMIC-LR and MNIST-CNN respectively. 

\noindent\textbf{Results and Discussion}

\begin{figure}[tb!]
    \centering
    \subfigure[]{\includegraphics[width=0.24\textwidth]{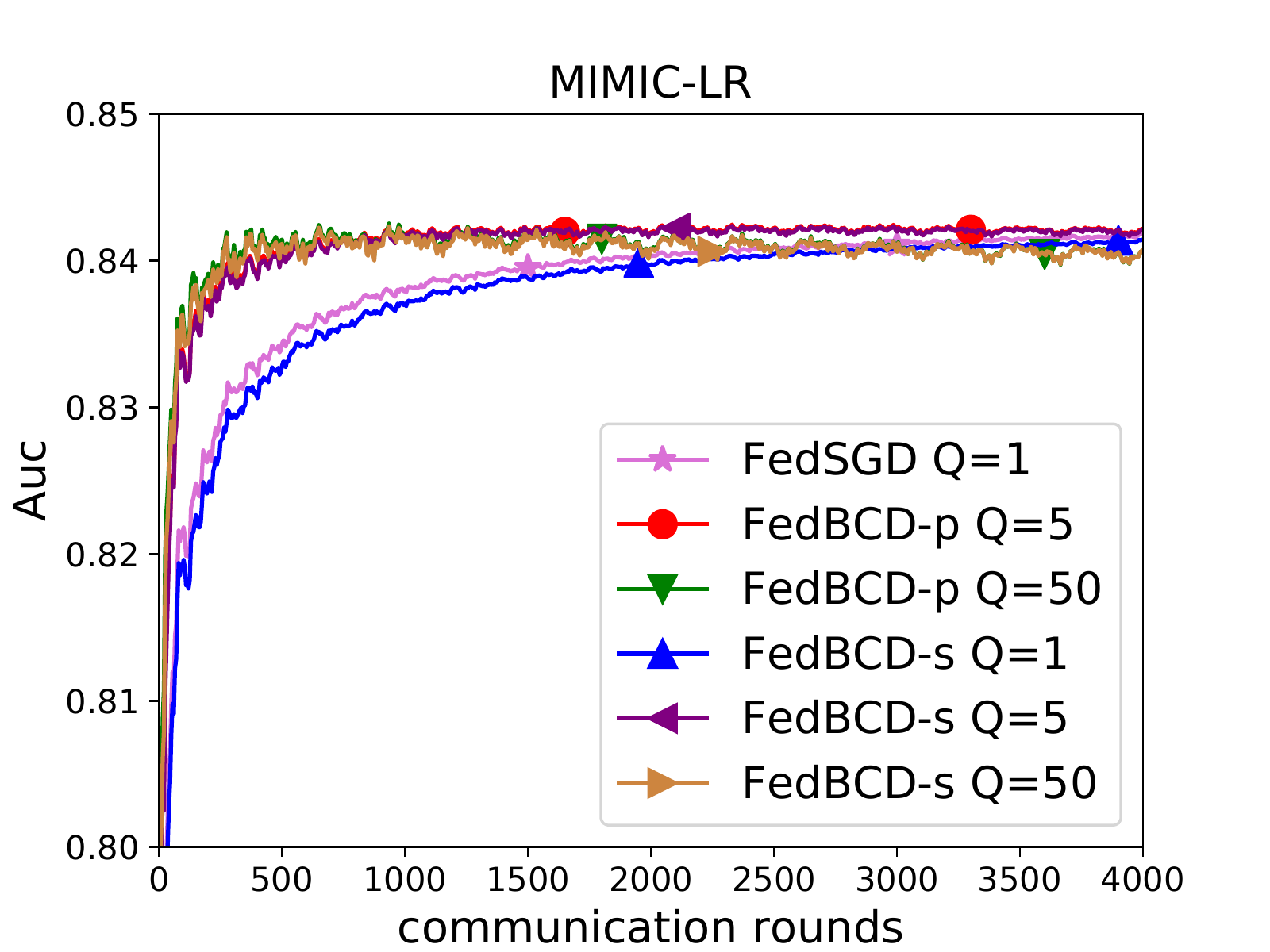}} 
    \subfigure[]{\includegraphics[width=0.24\textwidth]{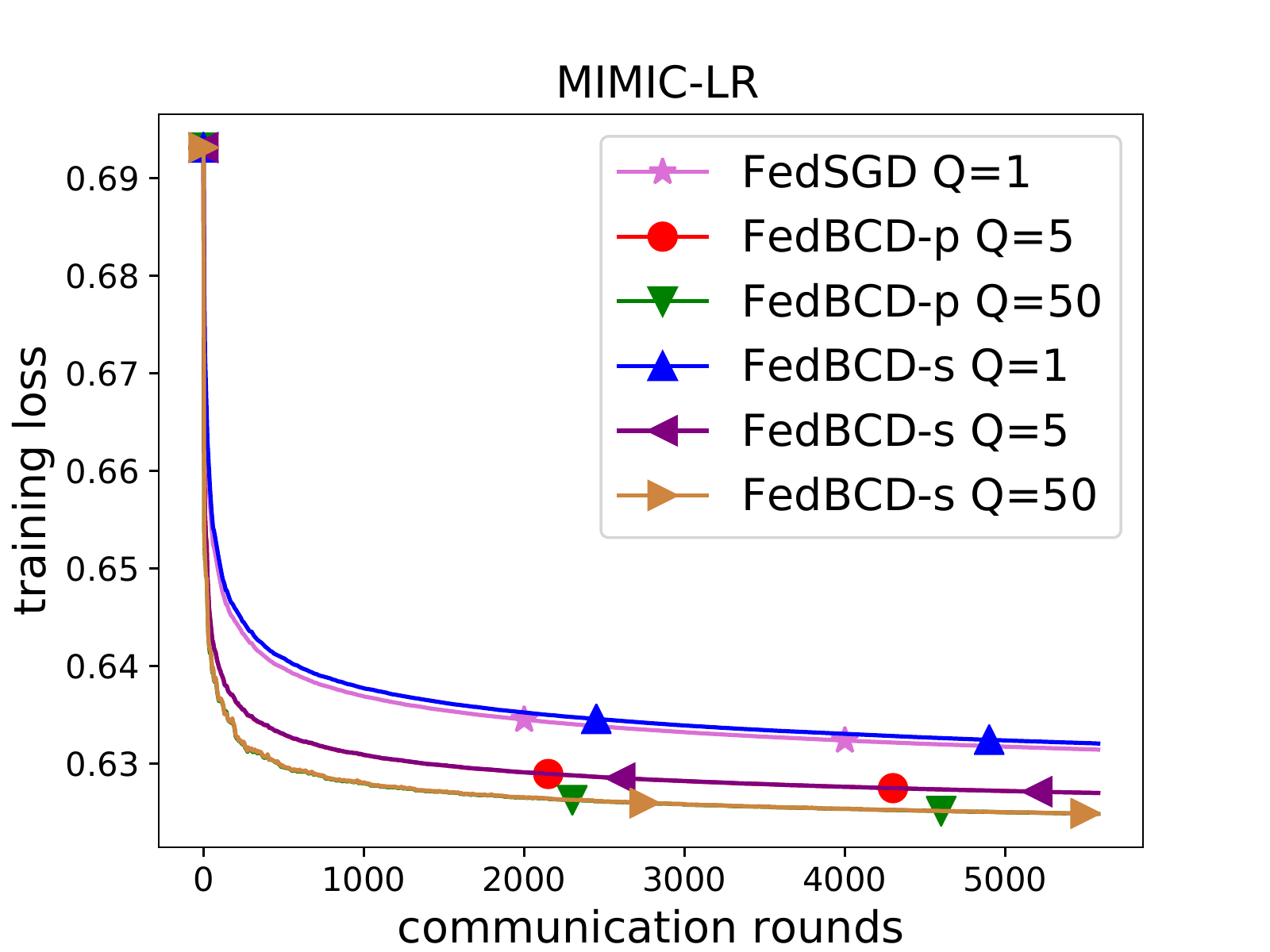}} 
    \subfigure[]{\includegraphics[width=0.24\textwidth]{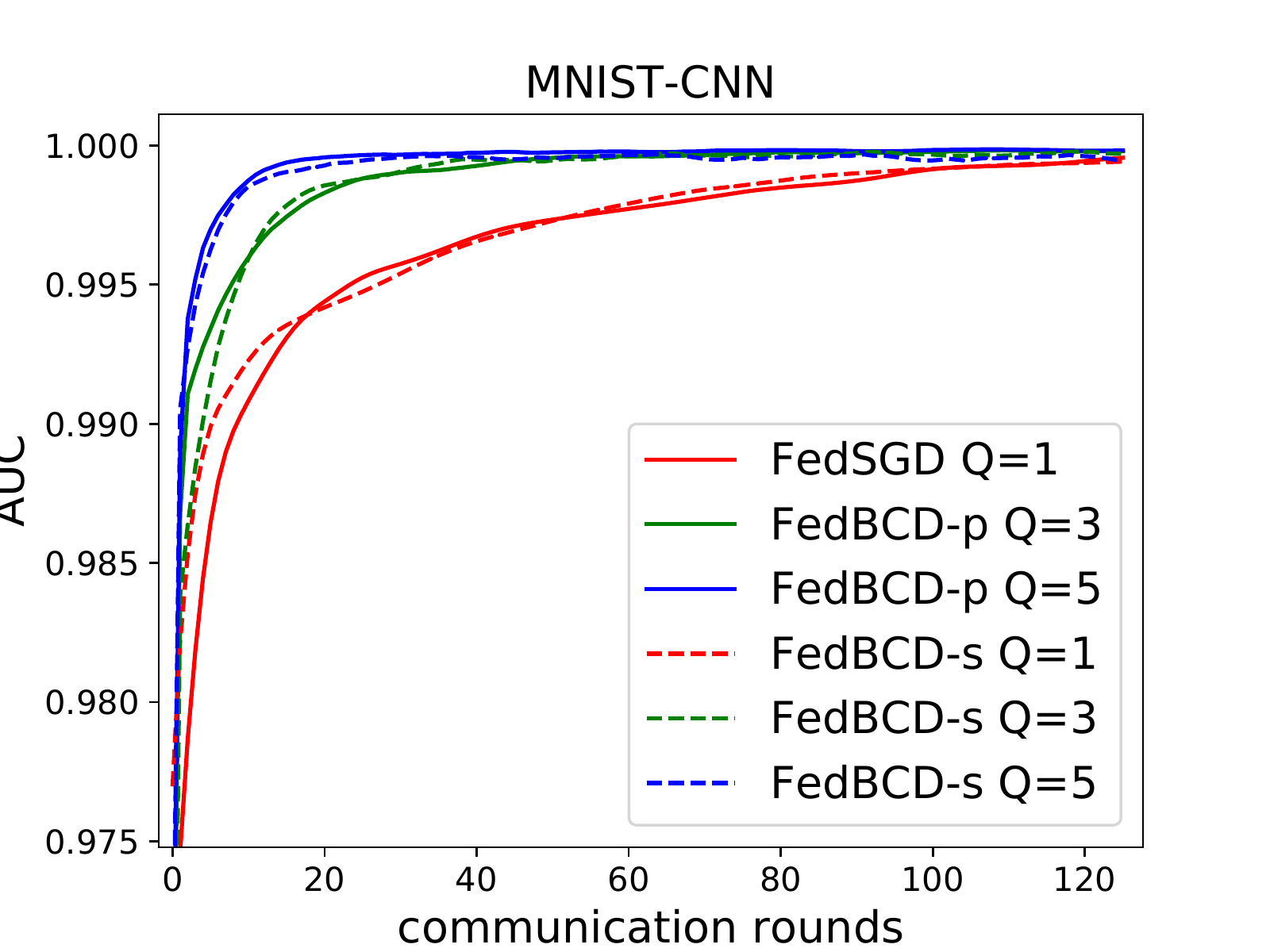}}
    \subfigure[]{\includegraphics[width=0.24\textwidth]{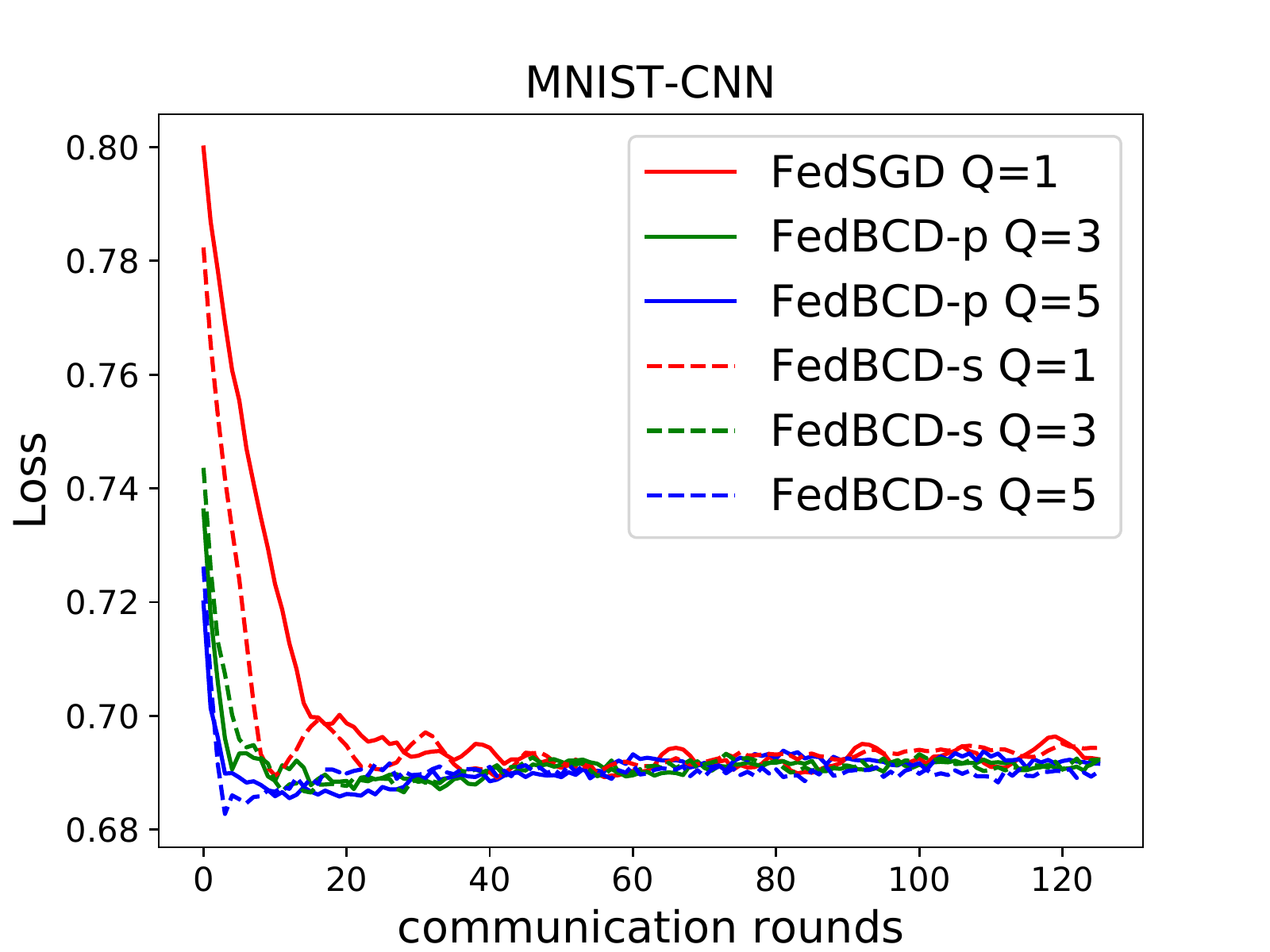}}
    \caption{Comparison of AUC (Left) and training loss (Right) in MIMIC-III dataset with varying local iterations, denoted by $Q$. }
    \label{varying_local_iter}
\end{figure}

\noindent\textbf{FedBCD-p vs FedBCD-s.}
We first study the impact of varying local iterations on the communication efficiency of both FedBCD-p and FedBCD-s algorithms based on MIMIC-LR and MNIST-CNN (Figure \ref{varying_local_iter}). We observe similar convergence for FedBCD-s and FedBCD-p for various values of $Q$.  
However, for the same communication round, the running time of FedBCD-s doubles that of FedBCD-p due to sequential execution. As the number of local iteration increases, we observe that the required number of communication rounds reduce dramatically (Table \ref{tab:1}). Therefore, by reasonably increasing the number of local iteration, we can take advantage of the parallelism on participants and save the overall communication costs by reducing the number of total communication rounds required.

\begin{figure}[h]
    \centering
    \includegraphics[width=4.1cm,height=3.5cm]{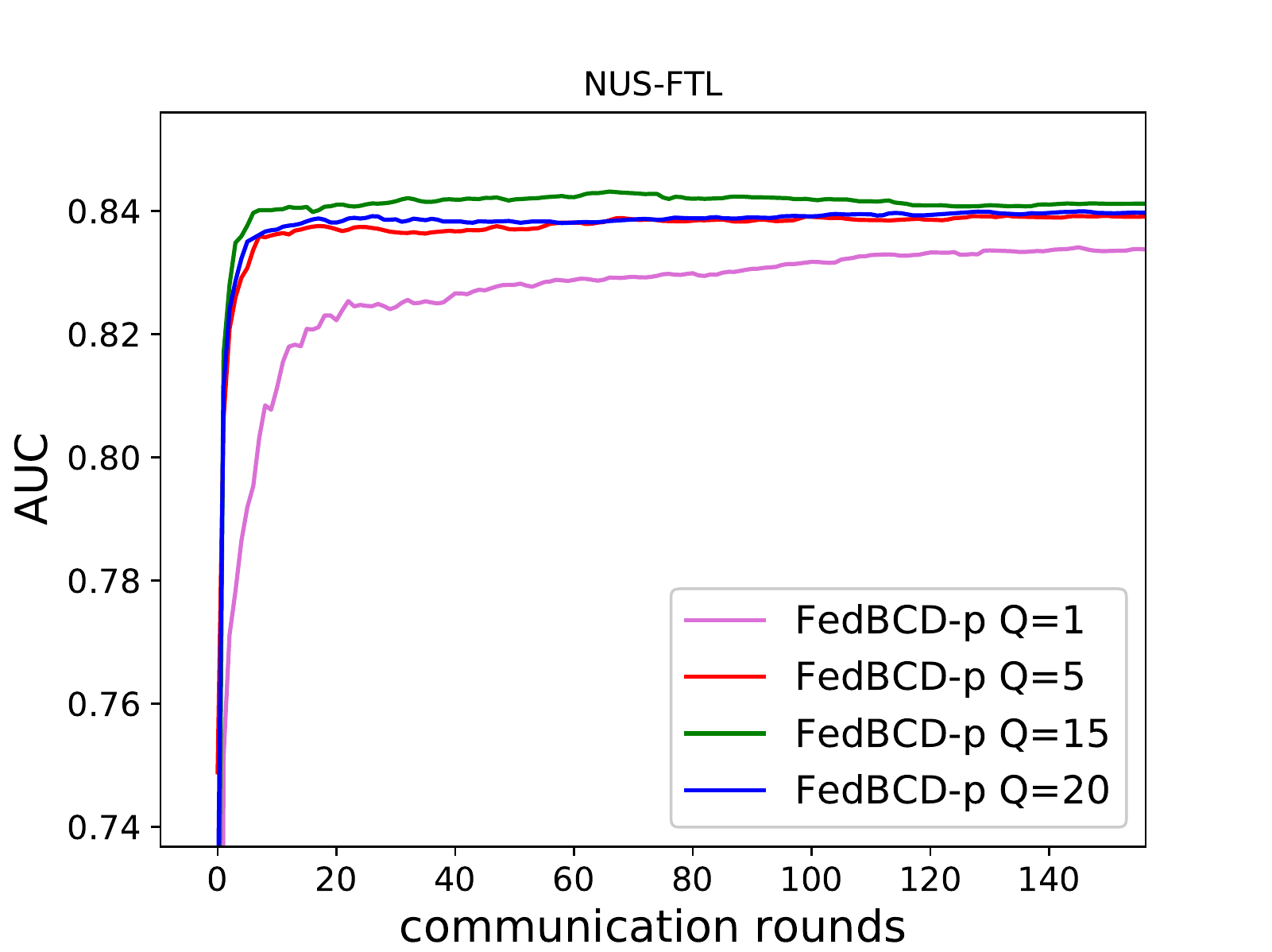}
    \includegraphics[width=4.1cm,height=3.5cm]{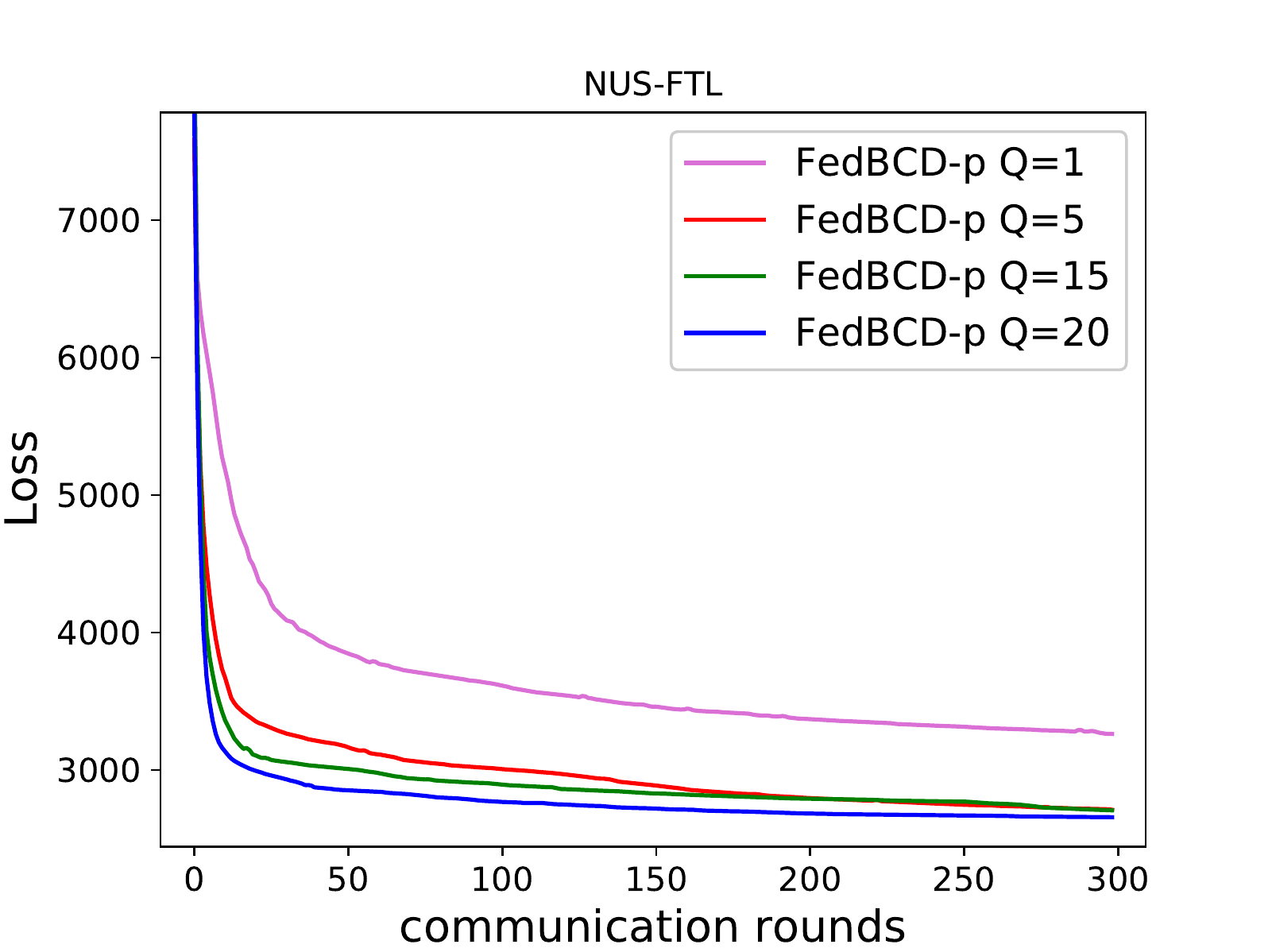}
    \caption{Comparison of AUC (Left) and training loss (Right) in NUS-WIDE dataset with varying local iterations, denoted by $L$. }
    \label{ftl_nus_wide}
\end{figure}

\begin{figure}[h]
    \centering
    \subfigure[]{\includegraphics[width=4.1cm,height=3.5cm]{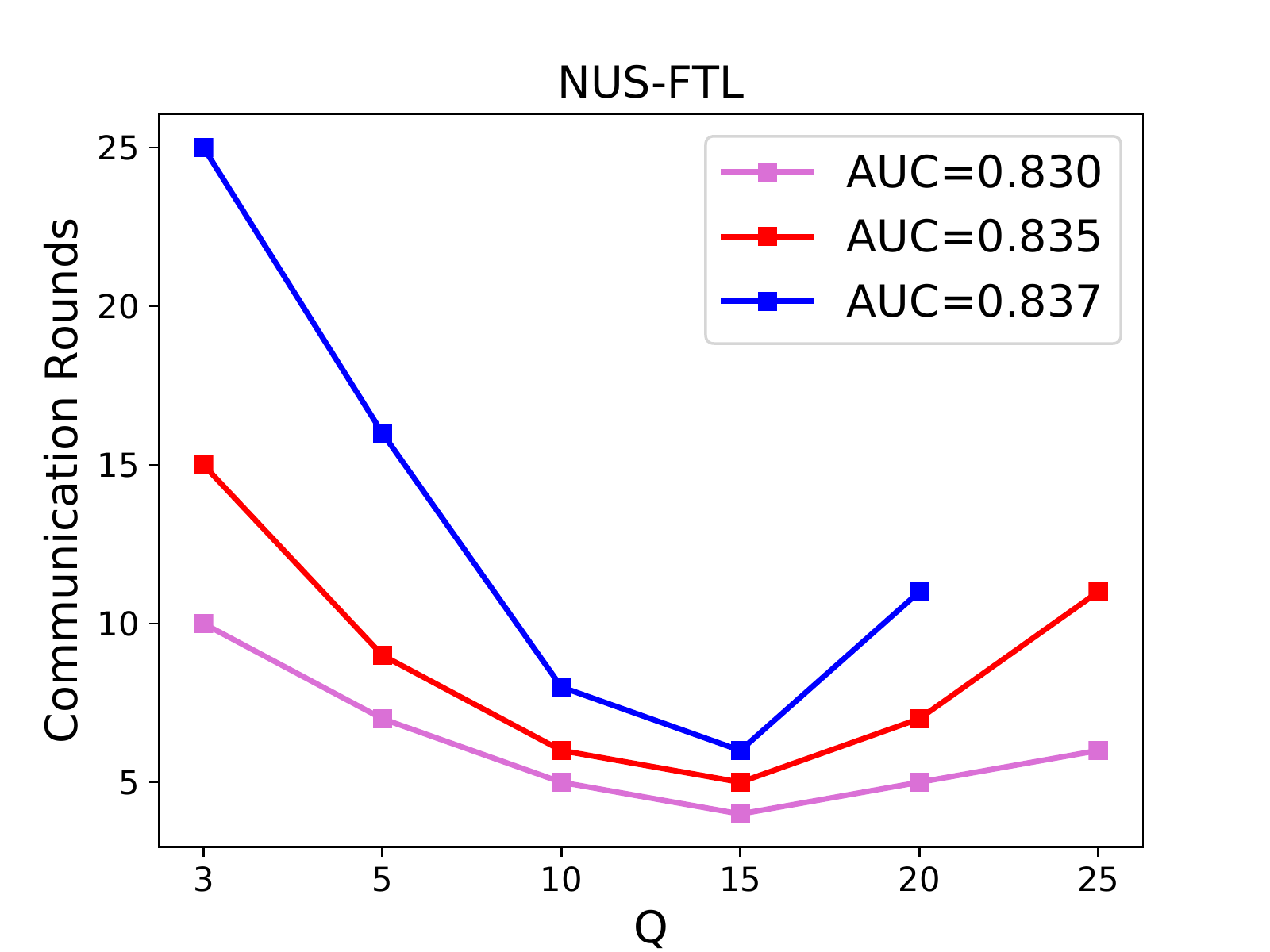} \label{Q_curve}}
    \subfigure[]{\includegraphics[width=4.1cm,height=3.5cm]{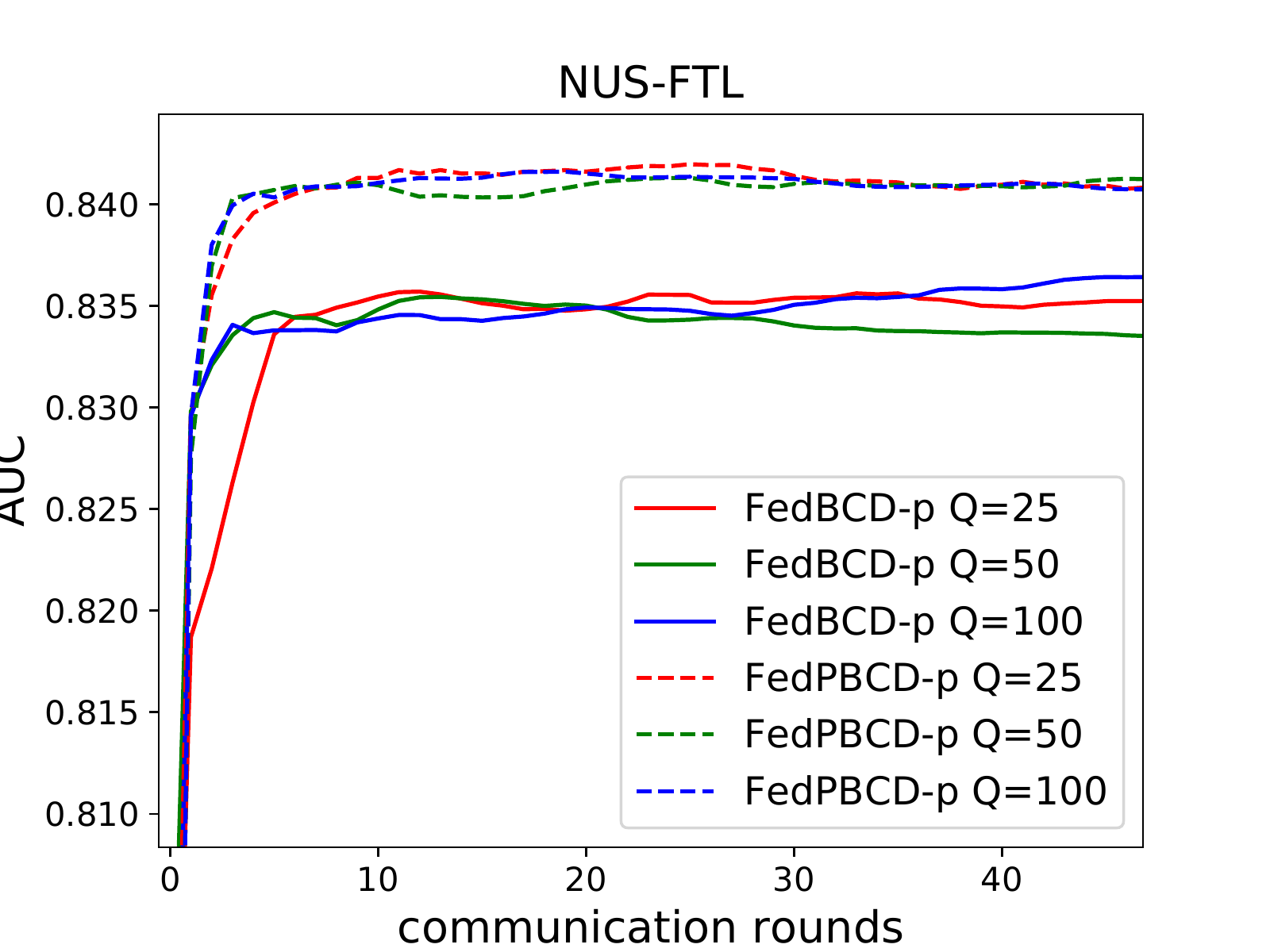} \label{large_Q}}
    \caption{The relationship between communication rounds and varying local iterations denoted by $Q$ for three target AUC (a), and the comparison between FedBCD-p and FedPBCD-p for large local iterations (b). }
    \label{Q_curve_n_large_Q}
\end{figure}

\begin{table}[tb!]
    \centering
\begin{tabular}{c||c|c||c|c}
\hline
\multicolumn{1}{c}{} & \multicolumn{2}{c}{MIMIC-LR} & \multicolumn{2}{c}{MNIST-CNN} \\
\multicolumn{1}{c}{} & \multicolumn{2}{c}{AUC 84\%} & \multicolumn{2}{c}{AUC 99.7\%} \\
\multicolumn{1}{c}{Algo.} & \multicolumn{1}{c}{Q} & \multicolumn{1}{c}{rounds} & \multicolumn{1}{c}{Q}&\multicolumn{1}{c}{rounds} \\
\hline
FedSGD & 1 & 334 & 1 & 46 \\
\hline
FedBCD-p & 5 & 71 & 3 & 16 \\
  & 50 & 52 & 5 & 8 \\
\hline
FedBCD-s & 1 & 407 & 1 & 48 \\
  & 5 & 74 & 3 & 15 \\
  & 50 & 52 & 5 & 9 \\
\hline
\end{tabular}
    \caption{Number of communication rounds to reach a target AUC-ROC for FedBCD-p, FedBCD-s and FedSGD on MIMIC-LR and MNIST-CNN respectively.}
    \label{tab:1}
\end{table}
\noindent\textbf{Impact of $Q$.}
Theorem \ref{th:main} suggests that as $Q$ grows the required number of communication rounds may first decrease and then increase again, and eventually the algorithm may not converge to optimal solution. To further investigate the relationship between the convergence rate and the local iteration $Q$, we evaluate FedBCD-p algorithm on NUS-FTL with a large range of $Q$. The results are shown in Figure \ref{ftl_nus_wide} and Figure \ref{Q_curve}, which illustrate that FedBCD-p achieves the best AUC with the least number of communication rounds when $Q = 15$. For each target AUC, there exists an optimal $Q$. This manifests that one needs to carefully select $Q$ to achieve the best communication efficiency, as suggested by Theorem \ref{th:main}. 

%

Figure \ref{large_Q} shows that for very large local iteration $Q = 25, 50 \text{ and } 100$, the FedBCD-p cannot converge to the AUC of $83.7\%$. This phenomenon is also supported by Theorem \ref{th:main}, where if $Q$ is too large the right hand side of \eqref{eq:thm} may not go to zero. Next we further address this issue by making the algorithm less sensitive in choosing $Q$.

\noindent\textbf{Proximal Gradient Descent.}
~\cite{TianLi2019} proposed adding a proximal term to the local objective function to alleviate potential divergence when local iteration is large. Here, we explore this idea to our scenario. We rewrite \eqref{ca_local_grad} as follows:
\begin{align}
   g_k(\by^r_k;\mD_i)= g_k([\bTheta^{r_0}_{-k},\theta^r_k];\mD_i]) + {\mu}(\theta^r_k - \theta_k^{r_0})
\end{align}

where ${\mu}(\theta^r_k - \theta_k^{r_0})$ is the gradient of the proximal term $\frac{\mu}{2}||\theta^r_k - \theta_k^{r_0}||^2$, 
which exploits the initial model $\theta_k^{r_0}$ of party $k$ to limit the impact of local updates by restricting the locally updated model to be close to $\theta_k^{r_0}$. We denote the proximal version of FedBCD-p as FedPBCD-p. We then apply FedPBCD-p with $\mu=0.1$ to NUS-FTL for $Q=25$, 50 and 100 respectively. Figure \ref{large_Q} illustrates that if Q is too large, FedBCD-p fails to converge to optimal solutions whereas the FedPBCD-p converges faster and is able to reach at a higher test AUC than corresponding FedBCD-p does. 

\begin{figure}[h]
    \centering
    \includegraphics[width=4.1cm,height=3.5cm]{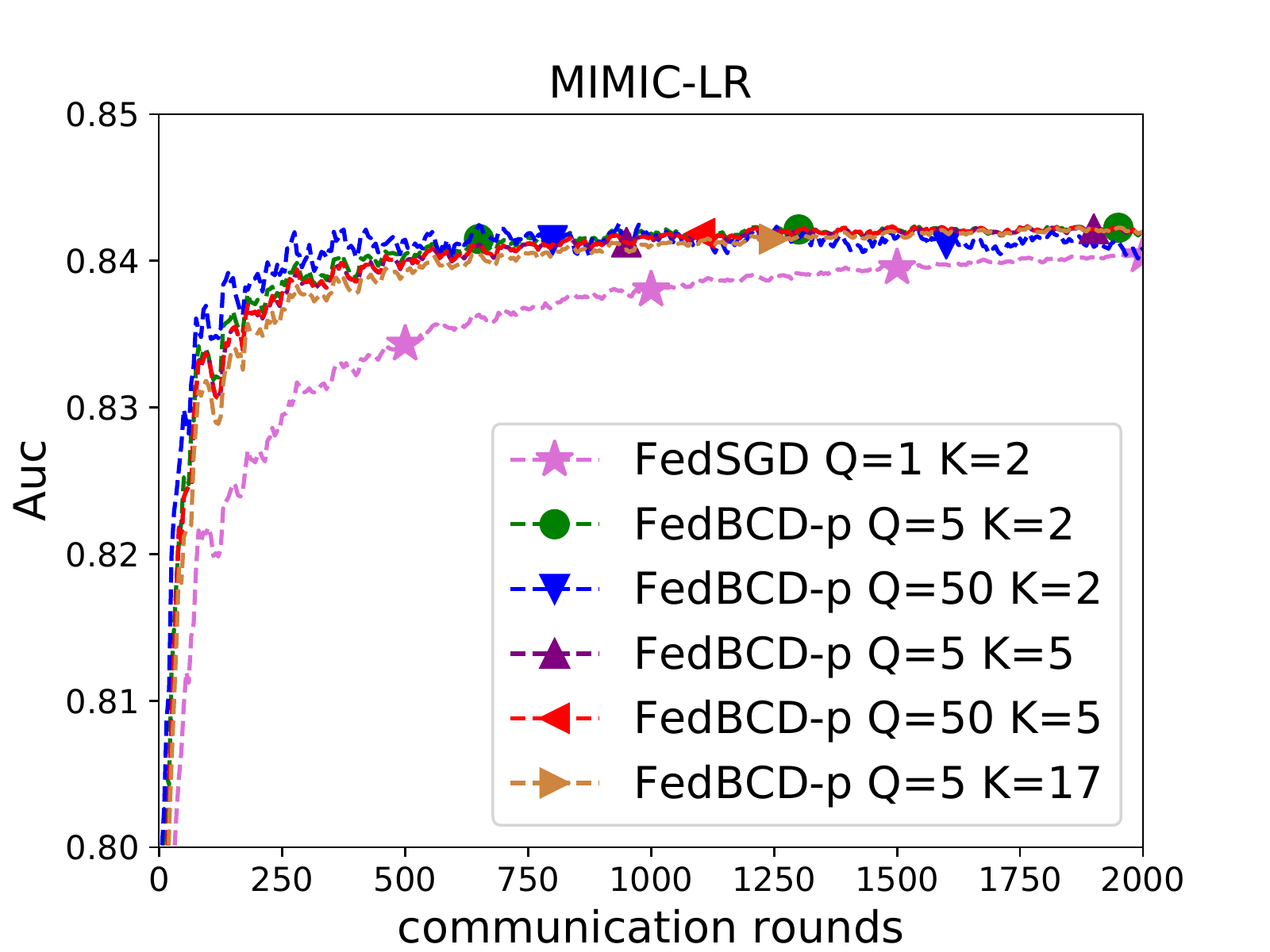}
    \includegraphics[width=4.1cm,height=3.5cm]{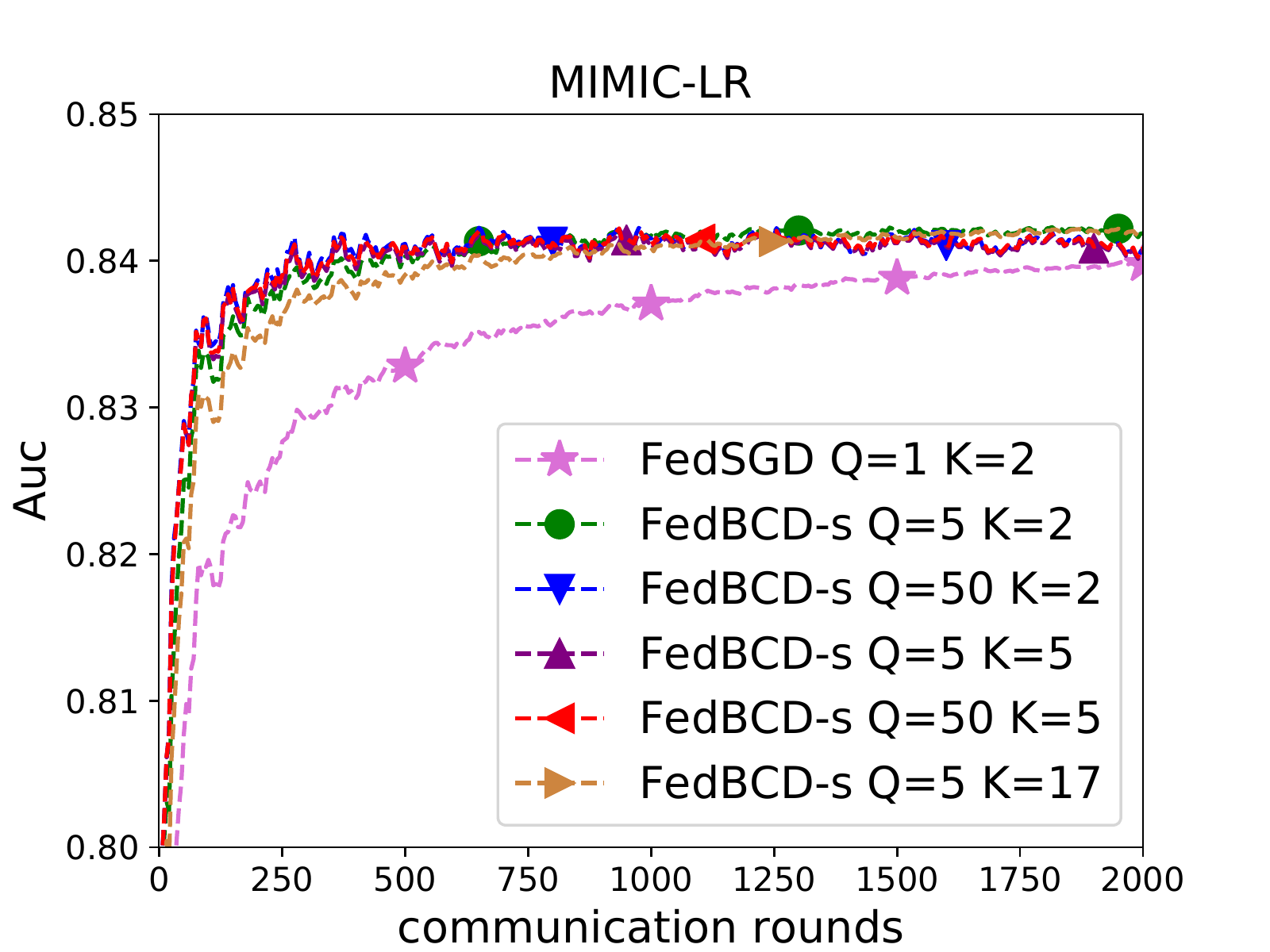}
    \caption{Comparison of AUC in MIMIC-III dataset with varying local iterations (denoted by $Q$) and number of parties (denoted by $K$). (Left) FedBCD-p; (Right) FedBCD-s}
    \label{lr-5}
\end{figure}

\noindent\textbf{Increasing number of Parties.}
In this section, we increase the number of parties to five and seventeen and conduct experiments for MIMIC-LR task. We partition data by clinical variables with each party having all the related features of the same variable.  
We adopt a decay learning rate strategy with $\frac{\eta_0}{\sqrt{Tk}}$ according to Theorem \ref{th:main}. The results are shown in Figure \ref{lr-5}. We can see that the proposed method still performs well when we increase the local iterations for multiple parties. As we increase the number of parties to five and seventeen, FedBCD-p is slightly slower than the two-party case, but the impact of node $K$ is very mild, which verifies the theoretical analysis in Remark 3. 

\noindent\textbf{Implementation with HE.} In this section, we investigate the efficiency of FedBCD-p algorithm with homomorphic encryption (HE) applied. Using HE to protect transmitted information ensures higher security but it is extremely computationally expensive to perform computations on encrypted data. In such a scenario, carefully selecting Q may reduce communication rounds but may also introduce computational overhead because the total number of local iterations may increase ($Q \times$ number of communication rounds). We integrated the FedBCD-p algorithm into the current FTL implementation on FATE\footnote{https://github.com/FederatedAI/FATE} and simulate two-party learning on two machines with Intel Xeon Gold model with 20 cores, 80G memory and 1T hard disk.
The experimental results are summarized in Table \ref{ftl_fate_he}. It shows that FedBCD-p with larger Q costs less communication rounds and total training time to reach a specific AUC with a mild increase in computation time but more than 70 percents reduction in communication round from FedSGD to $Q=10$. 



\begin{table}[h]
\footnotesize
\centering
\begin{tabular}{c|c||c||c|c|c|c}
\hline
\multicolumn{7}{c}{Credit-FTL} \\\hline
\multicolumn{1}{c}{AUC} & \multicolumn{1}{c}{Algo.} & \multicolumn{1}{c}{Q} & \multicolumn{1}{c}{R} &\multicolumn{1}{c}{comp.} & \multicolumn{1}{c}{comm.} & \multicolumn{1}{c}{total}\\
\hline
\hline
& FedSGD & 1 & 17 & 11.33 & 11.34 & 22.67 \\
$70\%$ & FedBCD-p & 5 & 4 & 13.40 & 2.94 & 16.34 \\ 
    &    & 10 & 2 & 10.87 & 2.74 & 13.61 \\
\hline
\hline
& FedSGD & 1 & 30 & 20.50 & 20.10 & 40.60\\
$75\%$& FedBCD-p 
      & 5 & 8 & 26.78 & 5.57 & 32.35 \\
      &   & 10 & 4 & 23.73 & 2.93 & 26.66 \\
\hline
\hline
& FedSGD & 1 & 46 & 32.20  & 30.69 &  62.89\\
$80\%$ & FedBCD-p 
         & 5 & 13 & 43.52 & 9.05 & 52.57 \\
    &     & 10 & 7 & 41.53 & 5.12 & 46.65 \\

\hline
\end{tabular}
    \caption{Number of communication rounds and training time to reach target AUC $70\%$, $75\%$ and $80\%$ respectively for FedSGD versus FedBCD-p. R, comp. and comm. denote communication rounds, computation time (mins), and communication time (mins) respectively.}
    \label{ftl_fate_he}
\end{table}

\vspace{-5pt}

\section{Conclusions and Future Work}\label{sec:summary}
In this paper, we propose a new collaboratively learning framework for distributed features based on block coordinate gradient descent in which parties perform more than one local update of gradients before communication. Our approach significantly reduces the number of communication rounds and the total communication overhead. We theoretically prove that the algorithm achieves global convergence with a decay learning rate and proper choice of $Q$. 
The approach is supported by our extensive experimental evaluations. We also show that adding proximal term can further enhance convergence at large value of $Q$.  
Future work may include investigating and further improving the communication efficiency of such approaches for more complex and asynchronized collaborative systems. 

\clearpage

 \bibliographystyle{nips}
 \addcontentsline{toc}{chapter}{\protect\numberline{}{References}}
 \bibliography{neurips_2019}
 \newpage

\pagebreak

\textbf{\large Supplementary Materials}

\appendix
\section{Proof of Theorem 1}
\begin{lemma} \label{lemma:s1} For any vector $\theta^0 \in R^{d_k}$. There exists infinite many of non-identity orthogonal matrix $U \in R^{d_k\times d_k}$ such that
\begin{eqnarray}
UU^T &=& I \label{U:prop:1} \\
U\theta^0 &=& \theta^0 \label{U:prop:2}
\end{eqnarray}
\end{lemma}
\noindent \emph{Proof.} {
First we construct a orthogonal $U_1$ satisfying \eqref{U:prop:1} and \eqref{U:prop:2} for 
\begin{equation}
    \theta^0 := e_1 = (1,0,\cdots, 0)^T \in R^{d_k}
\end{equation}
Then we complete the proof by generalizing the construction for an arbitrary $\theta^0 \in R^{d_k}$.

With $\theta^0 = e_1$, we construct $U_1$ in the following way
\begin{equation}
    U_1 := \left[ \begin{array}{c|c}
    1 & 0 \\ \hline 
    0 & V 
    \end{array}\right] \label{def:U1}
\end{equation}
where $V\in R^{(d_k-1)\times (d_k-1)}$ is any non-identity orthogonal matrix with $d_k>2$, i.e.,
\begin{equation}
    VV^T = I.
\end{equation}
Condition \eqref{U:prop:1} is satisfied since 
\begin{eqnarray}
    U_1U_1^T &=& \left[ \begin{array}{c|c}
    1 & 0 \\ \hline 
    0 & V 
    \end{array}\right] \left[ \begin{array}{c|c}
    1 & 0 \\ \hline 
    0 & V^T 
    \end{array}\right] \\
    &=& \left[ \begin{array}{c|c}
    1 & 0 \\ \hline 
    0 & VV^T 
    \end{array}\right] \\
    &=& I,
\end{eqnarray}

and condition \eqref{U:prop:2} is satisfied trivially, i.e.,
\begin{equation}
    U_1e_1 = \left[\begin{array}{c|c}
    1 & 0, \cdots, 0 
    \end{array}\right]^T = e_1
\end{equation}
For any arbitrary $\theta^0$, we apply the \textit{Householder transformation} to "rotate" it to the basis vector $e_1$, i.e.,
\begin{equation}
    \theta^0 = \|\theta^0\|_2Pe_1 \label{def:w0}
\end{equation}
where $P$ is the \textit{Householder transformation} operator such as
\begin{eqnarray}
P &=& P^T \\
PP^T &=& PP = I
\end{eqnarray}
Therefore from $U_1$ defined in (\ref{def:U1}) we can construct $U$ by
\begin{equation}
    U = PU_1P.
\end{equation}
Finally, we verifies that $U$ satisfies condition \eqref{U:prop:1}) and \eqref{U:prop:2}):
\begin{eqnarray}
UU^T &=& PU_1P(PU_1^TP) \\
&=& PU_1(PP)U_1^TP \\
&=& PU_1U_1^TP \\
&=& PP = I \\
U\theta^0 &=& PU_1P\theta^0 \\
&=& \|\theta^0\|_2P(e_1U_1) \label{eq:2}\\
&=& \|\theta^0\|_2Pe_1 \\
&=& \theta^0
\end{eqnarray}
where (\ref{eq:2}) holds since from (\ref{def:w0}) we have
\begin{equation}
    P\theta^0 = \|\theta^0\|_2e_1
\end{equation}
}
\subsection{Proof of Theorem 1}
\noindent \textbf{Proof:} We first show that the conclusion holds for the case when $k<K$. With initial weight $\theta_k^0 \in R^{d_k}$, Lemma \ref{lemma:s1} shows that we can find infinite number of non-identity matrix $U\in R^{d_k\times d_k}$ such that
\begin{equation}
    \theta_k^0 = U^T\theta_k^0 \label{cond:theta0_U}.
\end{equation}
Let $x_{i_j}^k$ denotes the $i$th sample of the data set $\mS_j$ sampled at $j$th iteration. We show that for any $\{x_{i_j}^k\}_{i \in \mS_j,j=0,1,\cdots}$ that yields observations $\{H^k_{i_j}\}_{j=0,1,\cdots}$, we can construct another set of data
\begin{equation}
    \tilde{x}_{i_j}^k := x_{i_j}^kU \label{def:tilde_U}
\end{equation}
where $U$ is chosen to satisfy condition \eqref{cond:theta0_U}. Let  $\{\tilde{H}^k_{i_j}\}$ be observations generated by $\{\tilde{x}_{i_j}^k\}$, and $\{\tilde{\theta}_k^j\}$ be weight variables with
\begin{equation}
\tilde{\theta}_k^0 = U^T\theta_k^0. \label{def:theta0}
\end{equation}
We show in the following that for $j = 0, 1, \cdots$
\begin{eqnarray}
    \tilde{H}^k_{i_j} &=& H^k_{i_j}\label{cond:uA} \\
    \tilde{\theta}_k^j &=& U^T\theta_k^j. \label{cond:theta}
\end{eqnarray}
It is easy to verify \eqref{cond:uA} for $j=0$, i.e.,
\begin{eqnarray*}
H^k_{i_0} &=& x_{i_0}^kUU^T\theta_k^0  \\
&=& (x_{i_0}^kU)(U^T\theta_k^0)  \\
&=& \tilde{x}_{i_0}^k\tilde{\theta}_k^0  \\
&=& \tilde{H}^k_{i_0},
\end{eqnarray*}
From equation (11), we define
\begin{equation}
    g_{i_j} := g(H_{i_j},y^K_i)
\end{equation}
Now assuming that condition \eqref{cond:uA} and \eqref{cond:theta} hold for $j\leq J$. Then
\begin{eqnarray}
    g_{i_j} &=& \tilde{g}_{i_j} \label{induction:l}\\
    \tilde{\theta}_k^j &=& U^T\theta_k^j \label{induction:theta}
\end{eqnarray}
We first show \eqref{cond:theta} holds for $j=J+1$.
\begin{eqnarray}
&&\tilde{\theta}_k^{J+1} \\
&=& \tilde{\theta}_k^{J} - \eta(\frac{1}{N_{\mS_J}}\sum_{i \in \mS_J}\tilde{g}_{i_J}(\tilde{x}_{i_J}^k)^T + \lambda \tilde{\theta}_k^J) \\ 
&=& U^T\theta_k^J - \eta(\frac{1}{N_{\mS_J}}\sum_{i \in \mS_J}g_{i_J}(x_{i_J}^kU)^T + \lambda U^T\theta_k^J) \label{eq:mid:4}\\
&=& U^T(\theta_k^J - \eta(\frac{1}{N_{\mS_J}}\sum_{i \in \mS_J}g_{i_J}(x_{i_J}^k)^T + \lambda \theta_k^J)) \\
&=& U^T\theta_k^{J+1}
\end{eqnarray}
where \eqref{eq:mid:4} follows from \eqref{induction:l} and \eqref{induction:theta}. 
Note if $Q$ local updates are performed, locally we have
\begin{eqnarray}
\theta_k^{j,q+1} - \theta_k^{j,q} = (1-\eta\lambda)(\theta_k^{j,q} - \theta_k^{j,q-1})
\end{eqnarray}
where $\theta_k^{j,q}$ denotes the $q$th local update of $j$th iteration.
it is thus easy to show that
\begin{eqnarray}
\tilde{\theta}_k^{J+1,q} = U^T\theta_k^{J+1,q}
\end{eqnarray}
Next we show \eqref{cond:uA} holds for $j=J+1$.
\begin{eqnarray}
\tilde{H}^k_{i_{J+1}} &=& \tilde{x}^k_{i,J+1}\tilde{\theta}_k^{J+1} \\
&=& x^k_{i,J+1}UU^T\theta_k^{J+1} \\
&=& x^k_{i,J+1}\theta_k \\
&=& H^k_{i_{J+1}}
\end{eqnarray}
Finally we complete the proof by showing the conclusion holds for $k=K$. Similarly for any $\{x_{i_j}^K\}_{j=0,1,\cdots}$ and $\{y_{i_j}^K\}$, we construct a different solution as follows:
\begin{eqnarray}
    \tilde{x}_{i_j}^K &:=& x_{i_j}^KU \label{def:xb_tilde} \\
    \tilde{y}_{i_j}^K &:=& y_{i_j}^K. \label{def:yb_tilde} 
\end{eqnarray}
where $U \in R^{d_k\times d_k}$ is a non-identity orthogonal matrix satisfying \eqref{def:tilde_U} for $k=K$. Therefore we have
\begin{eqnarray}
H^K_{i_j} &=& g(H_{i_j}, y^K_{i_j}) \\
&=& g(\tilde{H}_{i_j}, \tilde{y}^K_{i_j}) \\
&=& \tilde{H}^K_{i_j}
\end{eqnarray}

\section{Proof of Theorem 2}
To begin with, let us first introduce some notations. 

\noindent{\bf Notations.} For notation simplicity, we divide the total data set $[N]$ into $\mS_1,\dots,\mS_B$, each with $\vert\mS_i\vert=S$. Let $r$ denote the iteration index, where each iteration one round of {\it local} update is performed among all the nodes; Let $$\bTheta^r={[{\btheta^{r}_{1}};\dots; {\btheta^{r}_{K}}]}\bydef{[{\by^{r}_{1,1}};\dots;{\by^{r}_{K,K}}]}$$ 
denotes the updated parameters at each iteration of each node. Note that $\{\bTheta^r\}$ is only a sequence of \textquotedblleft virtual" variable that helps the proof, but it is never explicitly formed in the algorithm. 

Let $r_0$ denote the latest iteration before $r$ with global communication, so $\bTheta^{r_0}=\by^{r_0}_k,~k=1,\dots,K$ and  $$\by^r_k={[\bTheta^{r_0}_{-k},\theta^r_k]}.$$ Also we denote the full gradient of the loss function as $\nabla \mL(\bTheta)$. 
Further denote the stochastic gradient as $$\bG\bydef{[{g_1(\by_1;\mS)};\dots;{g_K(\by_K;\mS)}]}.$$ Using the above notation, the update rule becomes: $\bTheta^{r+1}=\bTheta^r-\gamma \bG^r.$

We also need the following important definition. Let us denote the variables updated in the inner loops, using mini-batch $\mS_l, l\in[B]$ as $\bTheta^r_{[l]}$. That is, we have the following: 
\begin{align*}
    \bTheta^{r_0+1}_{[l],k}&\bydef\bTheta^{r_0}_k-\eta g^r_k(\bTheta^{r_0};S_l),\\
    \bTheta^{r_0+\tau}_{[l],k}&\bydef\bTheta^{r_0+\tau-1}_{[l],k}-\eta g^r_k(\by^{r_0+\tau-1}_{[l],k};S_l),
\end{align*}
where $$\by^{r}_{[l],k}\bydef{[\bTheta^{r_0}_{-k},\theta^r_{[l],k}]}.$$
Further define 
\begin{align}
    \bar{\by}^r_k\bydef [\by^{r}_{[1],k}; \cdots; \by^{r}_{[B],k}].
\end{align}
Note that if $\mS_i$ is sampled at $r_0$, then the quantities $\bTheta^{r_0+\tau}_{[l],k}, \; l\ne i$ are {\it }not realized in the algorithm. They are introduced only for analytical purposes.

\begin{algorithm}[tb!]
	\caption{{\small Federated Stochastic BCD Algorithm}}
	\label{alg:FSBCD}
		{\bfseries Input:} $\bTheta^{(0)}, \eta, Q$\\
		{Initialize: $\by^0_k\bydef\bTheta^{(0)},~k=1,\dots,K$}\\
		\For{$r=0,1,\ldots$}{
        \If{$r~mod~Q=0$}{
	        do global communication \\
	        \qquad$\by^{r}_k\bydef\T{[\T{\by^{r}_{1,1}},\dots,\T{\by^{r}_{K,K}}]}$\\
	        Randomly sample a mini-batch $\mS_i$ for nodes $1,\dots,K$\\
	    }
		Calculate $g^r_k(\by^r_k; \mS_i)$ at each node using $\mS_i$\\
		{\For {$k=1,\cdots, K$}{ 
		Local updates\\
		$\by^{r+1}_{k,j}\bydef\begin{cases}
			\by^{r}_{k,j},j=1,\dots, K,j\neq k\\ 
			\by^{r}_{k,j}-\eta g^r_k(\by^r_k; \mS_i), j=k
			\end{cases}$}
		}	
        }
\end{algorithm}

By using these notations, Algorithm 2 can be written in the compact form as in Algorithm \ref{alg:FSBCD}. Further, the uniform sampling assumption A2 implies the following:

\noindent\textbf{A2':} {\bf Unbiased Gradient  with Bounded Variance}. Suppose Assumption A2 holds. Then we have the following: 
\begin{equation*}
\mbE[g^r_k(\by^r_k; \mS_i)\mid {\{\by^{t}_k\}_{t=1}^{r_0}}]=\frac{1}{B}\sum^B_{\ell=1}g^r_k(\by^r_{[\ell],k};\mS_\ell)\bydef \nabla_k\bar\mL(\bar\by^r_k)
\end{equation*}
where the expectation is taken on the choice of $i$ at iteration $r_0$, and conditioning all the past histories of the algorithm up to {iteration {$r_0$}}; Further, the following holds:
\begin{equation*}
\mbE[\lVert g^r_k(\by^r_k;\mS_i)-\nabla_k\bar\mL(\bar\by^r_k)\rVert^2\mid  \{\by^{t}_k\}_{t=1}^{r_0}]\leq {\frac{\sigma^2}{S}}
\end{equation*}
where the expectation is taken over the random selection of the data point at iteration $r_0$; $\sigma^2$ is the variance introduced by sampling from a single data point.  


To begin our proof, we first present a lemma that bounds the difference of the gradients evaluated at the global vector $\bTheta^r$ and the local vector $\by^r_k$. 
\begin{lemma}{\bf Bounded difference of gradient.}\label{lemma_1}
	Under Assumption A1, A2, the difference between the partial gradients evaluated at $\bTheta^r$  and $\by^r_k$ is bounded by the following:
	\begin{equation*}
	\small
	\begin{aligned}
	\mbE&\sum^{K}_{k=1}\lVert \nabla_k \mL(\bTheta^r)-g_k(\by^r_k;\mS)\rVert^2\leq (K+3)\eta^2\sum^K_{k=1}L^2_kQ^2\frac{\sigma^2}{S}\\
	&\qquad+\eta^2Q\sum^{K}_{k=1}(\sum^K_{j=1}L^2_j+3L^2_k)\sum^{r-1}_{\tau=r-Q}\mbE\lVert \nabla_k \mL(\by^\tau_k)\rVert^2+K\frac{\sigma^2}{S},
	\end{aligned}
	\end{equation*}
	where $\sigma$ is some constant related to 
	the variances of the sampling process. 
\end{lemma}
\subsection{Proof of Lemma \ref{lemma_1}}
\emph{Proof.} {First we take the expectation and apply assumption A1,A2
	\begin{equation}\label{eq:lemma1}
	\begin{aligned}
	    \mbE&[\lVert\nabla_k \mL(\bTheta^r)-g^r_k(\by^r_k;\mS_i)\rVert^2]\\
	    &=\mbE[\mbE_{\mS_i}[\lVert\nabla_k \mL(\bTheta^r_{[i]})-g^r_k(\by^r_k;\mS_i)\rVert^2\mid \bTheta^{r_0}]]\\
	    &\leq2\mbE[\mbE_{\mS_i}[\lVert\nabla_k \mL(\bTheta^r_{[i]})-\nabla_k\bar\mL(\bar\by^r_k)\rVert^2\mid \bTheta^{r_0}]\\
	    &\qquad+\mbE_{\mS_i}[\nabla_k\bar\mL(\bar\by^r_k)-g^r_k(\by^r_k;\mS_i)\rVert^2\mid \bTheta^{r_0}]]\\
	    &\leq2\mbE[L^2_k\mbE_{\mS_i}[\frac{1}{B}\sum^B_{l=1}\lVert\bTheta^r_{[i]}-\by^r_{[l],k}\rVert^2\mid \bTheta^{r_0}]\\
	    &\qquad+\mbE_{\mS_i}[\lVert\nabla_k\bar\mL(\bar\by^r_k)-g^r_k(\by^r_k;\mS_i)\rVert^2\mid \bTheta^{r_0}]]\\
	    &\le2\mbE\left[L^2_k\mbE_{\mS_i}[\frac{1}{B}\sum^B_{l=1}\lVert\bTheta^r_{[i]}-\by^r_{[l],k}\rVert^2\mid \bTheta^{r_0}]\right]+{\frac{2\sigma^2}{S}}
	\end{aligned}
	\end{equation}
	Note that
	\begin{equation}
	\small
	\begin{aligned}\label{eq:diff_x}
    \lVert\bTheta^r_{[i]}&-\by^r_{[l],k}\rVert^2=\sum^K_{j=1}\lVert \btheta^r_{[i],j}-\by^r_{[l],k,j} \rVert^2\\
	&=\sum_{j\neq k}\lVert \btheta^r_{[i],j}-\btheta^{r_0}_j \rVert^2+\lVert \btheta^r_{[i],k}-\btheta^{r}_{[l],k} \rVert^2\\
	&=\sum_{j\neq k}\lVert \btheta^{r_0}_j-\eta\sum^{r-1}_{\tau=r_0}g^\tau_j(\by^\tau_{[i],j}; \mS_i)-\btheta^{r_0}_j \rVert^2\\
	&\quad+\lVert \eta\sum^{r-1}_{\tau=r_0}g^\tau_k(\by^\tau_{[i],k}; \mS_i)-\eta\sum^{r-1}_{\tau=r_0}g^\tau_k(\by^\tau_{[l],k}; \mS_l)\rVert^2\\
	&\leq\eta^2\bigg[\sum_{j\neq k}\lVert \sum^{r-1}_{\tau=r_0}g^\tau_j(\by^\tau_{[i],j}; \mS_i)\rVert^2\\
	&\quad+2\lVert \sum^{r-1}_{\tau=r_0}g^\tau_k(\by^\tau_{[l],k}; \mS_l)\rVert^2+2\lVert \sum^{r-1}_{\tau=r_0}g^\tau_k(\by^\tau_{[l],k}; \mS_i)\rVert^2\bigg]\\
	&\leq \eta^2 (r-r_0-1)\sum^{r-1}_{\tau=r_0}(\sum^K_{j=1}\lVert g^\tau_j(\by^\tau_{[i],j}; \mS_i)\rVert^2\\
	&\quad+2\lVert g^\tau_k(\by^\tau_{[l],k}; \mS_l)\rVert^2+\lVert g^\tau_k(\by^\tau_{[i],k}; \mS_i)\rVert^2).
	\end{aligned}
	\end{equation}
	Notice that $r-r_0-1\leq Q$ and substitute \eqref{eq:diff_x} to \eqref{eq:lemma1}, by taking expectation over $\mS_i$, we obtain
	\begin{equation}
	\small
	\begin{aligned}
	\mbE&[\lVert\nabla_k \mL(\bTheta^r)-g^r_k(\by^r_k;\mS_i)\rVert^2]\\
	&\leq 2L^2_k\eta^2(r-r_0-1)\sum^{r-1}_{\tau=r_0}\mbE\bigg[\mbE_{\mS_i}\big[\frac{1}{B}\sum^B_{l=1}(\sum^K_{j=1}\lVert g^\tau_j(\by^\tau_{[i],j}; \mS_i)\rVert^2\\
	&\quad+2\lVert g^\tau_k(\by^\tau_{[l],k}; \mS_l)\rVert^2+\lVert g^\tau_k(\by^\tau_{[i],k}; \mS_i)\rVert^2)\big]\bigg]+{\frac{2\sigma^2}{S}}\\
	&\stackrel{(i)}\leq 2L^2_k\eta^2Q\sum^{r-1}_{\tau=r_0}\mbE\bigg[\mbE_{\mS_i}\big[(\sum^K_{j=1}\lVert g^\tau_j(\by^\tau_{[i],j}; \mS_i)\rVert^2\\
	&\quad+2\lVert \nabla_k\bar\mL(\bar\by^\tau_k)\rVert^2+{\frac{2\sigma^2}{S}}+\lVert g^\tau_k(\by^\tau_{[i],k}; \mS_i)\rVert^2)\big]\bigg]+{\frac{2\sigma^2}{S}}\\
	&\stackrel{(ii)}\leq 2L^2_k\eta^2Q\sum^{r-1}_{\tau=r_0}\mbE[(\sum^K_{j=1}(\lVert \nabla_k\bar\mL(\bar\by^\tau_k)\rVert^2+{\frac{\sigma^2}{S}})\\
	&\quad+3\lVert \nabla_k\bar\mL(\bar\by^\tau_k)\rVert^2+{\frac{3\sigma^2}{S}})]+{\frac{2\sigma^2}{S}}\\
	&\leq 2L^2_k\eta^2Q\sum^{r-1}_{\tau=r-Q}\mbE[(\sum^K_{j=1}\lVert \nabla_k\bar\mL(\bar\by^\tau_k)\rVert^2+{\frac{(K+3)\sigma^2}{S}}\\
	&\quad+3\lVert \nabla_k\bar\mL(\bar\by^\tau_k)\rVert^2)]+{\frac{2\sigma^2}{S}}\\
	\end{aligned}
	\end{equation}
	where in $(i)$ we used the definition of $\nabla_k\bar{\mathcal{L}}(\bar{\by}^r_k)$; in $(ii)$ we use the fact that $\mathbb{E}[x^2] -(\mathbb{E}[x])^2 = \mathbb{E}[(x-\mathbb{E}(x))^2]$; in  
	the last inequality is true because summing from $r_0$ to $r-1$ is always less that summing from $r-Q$ to $r-1$.
	Next we sum over K, we have
	\begin{equation}\label{eq:bound}
	\small
	\begin{aligned}
	\mbE&\sum^{K}_{k=1}\lVert \nabla_k \mL(\bTheta^r)-g_k(\by^r_k;\mS_i)\rVert^2\leq 2(K+3)\eta^2\sum^K_{k=1}L^2_kQ^2\frac{\sigma^2}{S}\\
	&\qquad+2\eta^2Q\sum^{K}_{k=1}(\sum^K_{j=1}L^2_j+3L^2_k)\sum^{r-1}_{\tau=r-Q}\mbE\lVert \nabla_k\bar\mL(\bar\by^\tau_k)\rVert^2+2K\frac{\sigma^2}{S},
	\end{aligned}
	\end{equation}
	This completed the proof of this result.
}
\subsection{Proof of Theorem 2}

\emph{Proof.}{
	First apply the Lipschitz condition of $\mL$, we have
	\begin{equation}\label{eq:main_0}
	\begin{aligned}
	\mL(\bTheta^{r+1})\leq&\mL(\bTheta^r)+\langle \nabla\mL(\bTheta^r),\bTheta^{r+1}-\bTheta^r\rangle\\
	&+\frac{L}{2}\lVert\bTheta^{r+1}-\bTheta^r\rVert^2.
	\end{aligned}
	\end{equation}
	Next we apply the update step in \algref{alg:FSBCD} and take the expectation
	\begin{equation}\label{eq:main_1}
	\small
	\begin{aligned}
	\mbE&[\mL(\bTheta^{r+1})-\mL(\bTheta^r)]\leq-\eta\mbE\langle \nabla\mL(\bTheta^r),\bG^r\rangle+\frac{L}{2}\eta^2\mbE\lVert\bG^r\rVert^2\\
	&=-\frac{\eta}{2}\mbE\lVert \nabla\mL(\bTheta^r)\rVert^2-\frac{\eta}{2}\mbE\lVert\bG^r\rVert^2\\
	&\qquad+\frac{\eta}{2}\mbE\lVert \nabla\mL(\bTheta^r)-\bG^r\rVert^2+\frac{L}{2}\eta^2\mbE\lVert\bG^r\rVert^2\\
	&=-\frac{\eta}{2}\mbE\lVert \nabla\mL(\bTheta^r)\rVert^2-\frac{\eta}{2}(1-L\eta)\mbE\lVert\bG^r\rVert^2\\
	&\qquad+\frac{\eta}{2}\mbE\sum^{K}_{k=1}\lVert \nabla_k \mL(\bTheta^r)-g_k(\by^r_k;\mS_i)\rVert^2
	\end{aligned}
	\end{equation}
    Also we have
	\begin{equation}\label{eq:main_3}
	\small
	\begin{aligned}
	\mbE\lVert\bG^r\rVert^2&=\mbE\lVert\mbE[\bG^r]\rVert^2+\mbE\lVert\bG^r-\mbE[\bG^r]\rVert^2\\
	&=\mbE\sum^{K}_{k=1}\lVert \nabla_k \bar\mL(\bar\by^r_k)\rVert^2+\mbE\sum^{K}_{k=1}\lVert\bG^r-\nabla_k \bar\mL(\bar\by^r_k)\rVert^2\\
	&\geq\mbE\sum^{K}_{k=1}\lVert \nabla_k \bar\mL(\bar\by^r_k)\rVert^2.
	\end{aligned}
	\end{equation}
	Substitute \eqref{eq:main_3} into \eqref{eq:main_1}, choose $0<\eta\leq\frac{1}{L}$ and apply \leref{lemma_1}, we have
	\begin{equation}\label{eq:main_4}
	\small
	\begin{aligned}
	\mbE&[\mL(\bTheta^{r+1})-\mL(\bTheta^r)]\\
	&\leq-\frac{\eta}{2}\mbE\lVert \nabla\mL(\bTheta^r)\rVert^2-\frac{\eta}{2}(1-L\eta)\mbE\sum^{K}_{k=1}\lVert \nabla_k \bar\mL(\bar\by^r_k)\rVert^2\\
	&\qquad+\frac{\eta}{2}\mbE\sum^{K}_{k=1}\lVert \nabla_k \mL(\bTheta^r)-g_k(\by^r_k;\mS_i)\rVert^2\\
	&\stackrel{\eqref{eq:bound}}\leq-\frac{\eta}{2}\mbE\lVert \nabla\mL(\bTheta^r)\rVert^2-\frac{\eta}{2}(1-L\eta)\mbE\sum^{K}_{k=1}\lVert \nabla_k \bar\mL(\bar\by^r_k)\rVert^2\\
	&\qquad+(K+3)\eta^3\sum^K_{k=1}L^2_kQ^2\frac{\sigma^2}{S}+\eta K\frac{\sigma^2}{S}\\
	&\qquad+\eta^3Q\sum^{K}_{k=1}(\sum^K_{j=1}L^2_j+3L^2_k)\sum^{r-1}_{\tau=r-Q}\mbE\lVert \nabla_k \bar\mL(\bar\by^\tau_k)\rVert^2\\
	\end{aligned}
	\end{equation}
	Average over $T$ and reorganize the terms, let $\eta$ satisfy $$1-L\eta-2\eta^2Q^2(\sum^K_{j=1}L^2_j+3L^2_k)\geq0,$$ ($0<\eta\leq \frac{\sqrt{2}}{2Q\sqrt{\sum^K_{j=1}L^2_j+3L^2_k}}$), then we obtain
	\begin{equation}\label{eq:main_5}
	\small
	\begin{aligned}
	\frac{1}{T}&\sum^{T-1}_{r=0}\mbE\lVert \nabla\mL(\bTheta^r)\rVert^2\leq\frac{2}{\eta T}\mbE[\mL(\bTheta^{0})-\mL(\bTheta^{T+1})]\\
	&\qquad-\sum^{K}_{k=1}(1-L\eta-2\eta^2Q^2(\sum^K_{j=1}L^2_j+3L^2_k))\mbE\lVert \nabla_k \bar\mL(\by^r_k)\rVert^2\\
	&\qquad+(K+3)2\eta^2\sum^K_{k=1}L^2_kQ^2\frac{\sigma^2}{S}+2K\frac{\sigma^2}{S}\\
	&\leq \frac{2}{\eta T}\mbE[\mL(\bTheta^{0})-\mL(\bTheta^{T+1})]\\
	&\qquad+(K+3)2\eta^2\sum^K_{k=1}L^2_kQ^2\frac{\sigma^2}{S}+2K\frac{\sigma^2}{S}.\\
	\end{aligned}
	\end{equation}
	The proof is completed.
}

\end{document}